\documentclass{article}
\usepackage{arxiv}

\usepackage[T1]{fontenc}
\usepackage{lmodern}          
\usepackage{amsmath,amssymb,amsfonts}
\usepackage{amsthm}            
\usepackage{mathtools}         
\usepackage{bm}                
\usepackage{graphicx}          
\usepackage{caption}
\usepackage{subcaption}        
\usepackage{booktabs}          
\usepackage{enumitem}          
\usepackage{geometry}          
\usepackage{hyperref}          
\hypersetup{
    colorlinks=true,
    linkcolor=cyan,
    citecolor=red,
    urlcolor=blue
}
\usepackage{doi}               
\usepackage{algorithm}
\usepackage{algpseudocode}     
\usepackage{afterpage}

\theoremstyle{definition}

\theoremstyle{remark}

\usepackage[normalem]{ulem}
\usepackage{tabularx}

\title{%
  {PIP$^2$ Net: Physics-informed Partition Penalty Deep Operator Network}
}
\author{%
    Hongjin Mi \\
  School of Mathematics\\ 
  Shanghai University of Finance and Economics\\ No.777 Guoding Road, Shanghai 200433, China\\
  \texttt{mhj6452021@163.com}
   \And
   Huiqiang Lun\\
   Faculty of Liberal Arts and Professional Studies \\
   York University\\
   4700 Keele St, North York, ON M3J1P3, Canada\\
   \texttt{brucehuiqianglun@gmail.com}
   \And
    Changhong Mou \\
  Department of Mathematics and Statistics\\ Utah State University \\
  900 Old Main Hill, Logan, UT 84322, USA \\
  \texttt{changhong.mou@usu.edu} 
  \And
    Yeyu Zhang \thanks{Corresponding author. Email: \texttt{zhangyeyu@mail.shufe.edu.cn}}\\
  School of Mathematics\\ 
  Shanghai University of Finance and Economics\\ No.777 Guoding Road, Shanghai 200433, China\\
  \texttt{zhangyeyu@mail.shufe.edu.cn} 
}
\date{\today}

\begin{document}
\maketitle
\begin{abstract}
Operator learning has become a powerful tool for accelerating the solution of parameterized partial differential equations (PDEs), enabling rapid prediction of full spatiotemporal fields for new initial conditions or forcing functions. Existing architectures such as DeepONet and the Fourier Neural Operator (FNO) show strong empirical performance but often require large training datasets, lack explicit physical structure, and may suffer from instability in their trunk-network features, where mode imbalance or collapse can hinder accurate operator approximation. Motivated by the stability and locality of classical partition-of-unity (PoU) methods, we investigate PoU-based regularization techniques for operator learning and develop a revised formulation of the existing POU--PI--DeepONet framework. The resulting \emph{P}hysics-\emph{i}nformed \emph{P}artition \emph{P}enalty Deep Operator Network (PIP$^{2}$ Net) introduces a simplified and more principled partition penalty that improved the coordinated trunk outputs that leads to more expressiveness without sacrificing the flexibility of DeepONet. We evaluate PIP$^{2}$ Net on three nonlinear PDEs: the viscous Burgers equation, the Allen--Cahn equation, and a diffusion--reaction system. The results show that it consistently outperforms DeepONet, PI-DeepONet, and POU-DeepONet in prediction accuracy and robustness.

\end{abstract}
\raggedbottom

\section{Introduction}
Many problems in engineering and physics are described by parameterized partial differential equations (PDEs) \cite{evans2022partial,farlow1993partial,courant1967partial}. In such problems, we often need to solve the same PDE many times, for different initial conditions, boundary conditions, source terms, or physical parameters \cite{berselli2006mathematics,hesthaven2016certified}. Standard numerical methods, such as finite difference, finite element, or spectral schemes, can be very accurate for a single solution, but they quickly become too expensive in multi-query settings like uncertainty quantification, inverse problems, design optimization, and real-time control \cite{layton2008introduction,hughes2003finite,leveque2007finite}. 
This motivates the use of operator learning \cite{li2021fourier,lu2021learning}, a framework that learns nonlinear mappings between infinite-dimensional function spaces, which further allows input data such as initial conditions to be mapped directly to the full PDE solution.
Operator learning has therefore become a powerful tool for accelerating PDE related simulations. Instead of predicting a single output for fixed inputs, operator learning  approximate the {solution operator} itself with the mapping of an given function e.g., an initial condition or forcing function to the corresponding spatiotemporal field of the governing equations \cite{kovachki2024operator,boulle2024mathematical}.

Operator learning can be broadly categorized into two different architectures: Deep Operator Network (DeepONet) and Fourier Neural Operator. Both tends to learn solution operators between infinite-dimensional function spaces, but they differ in how the mapping is represented and evaluated.
The Deep Operator Network (DeepONet), introduced by Lu et al.~\cite{lu2021learning} and supported by the universal operator approximation theorem of Chen and Chen~\cite{chen1995universal}, learns operators by splitting the representation into two components: a branch network that takes the input function, and a trunk network that includes spatial–temporal coordinates. The branch network outputs a set of coefficients derived from the input function, while the trunk network evaluates features at the desired coordinates. The final prediction is obtained by combining these two outputs. In this way, DeepONet can infer the entire solution field for new inputs that captures nonlinear operator mappings without requiring explicit spatial discretization of the underlying PDE.
On the other hand, the Fourier Neural Operator (FNO) adopts a spectral approach \cite{li2021fourier}. Rather than working in physical space, FNO learns the operator by applying trainable convolution kernels in Fourier space that enables efficient representation of global interactions across the domain. By working in the frequency domain, FNO can efficiently capture broad spatial influence and still scale well as resolution increases, since updates are performed with fast Fourier transforms instead of pointwise operations.
Purely data-driven operator networks, however, often require large quantities of high-fidelity training data and may generalize poorly outside the distribution they are trained on. A natural strategy to reduce this data burden is to incorporate physics information into the learning process. Physics-informed neural networks (PINNs)~\cite{raissi2019physics} introduce this idea by adding the governing PDE residual, along with initial and boundary conditions directly into the loss function through automatic differentiation. Building on this formulation, the Physics-Informed Deep Operator Network (PI-DeepONet)~\cite{wang2021learning} extends physics regularization to operator learning by adding the standard DeepONet data loss with physics residual terms. This physics informed mechanism potentially improves data efficiency and robustness, and has motivated a growing number of research on physics-informed neural operator networks.
Recent research has pushed PI-DeepONet forward in several important ways. 
Adaptive loss balancing approaches have been introduced to mitigate competition between data and physics residuals and improve training stability~\cite{wang2022respecting,lu2023nsga,lu2025evolutionary}.
Multiscale and domain-decomposed extensions~\cite{liu2021multiscale,chen2025deeponet,yang2025dd,mou2025pas} have also been shown to improve performance in problems with sharp gradients, heterogeneous media, or stiff dynamics.

A key difficulty underlying these challenges lies in how the operator is represented: specifically, in the features generated by the trunk network. In DeepONet, the trunk outputs act as coordinate-dependent modes that combine with branch coefficients to form the predicted solution field. When no structural guidance is imposed, these modes may become unbalanced: certain components grow disproportionately while others collapse toward zero. This imbalance can degrade conditioning, increase sensitivity to hyperparameters, and reduce interpretability of the learned operator.
In contrast, classical numerical methods often maintain stability through structured local representations. Among them, the \textit{partition-of-unity} (PoU) framework is especially notable: PoU basis constructions have long been used in meshfree and radial-basis-function (RBF) approximations~\cite{pou,rbf_pu_pde,rbf_pu_basket,rbf_pu_precond,rbf_pu_convection,SWJCP2018} to assemble globally smooth solutions from balanced local components. Motivated by these advantages, PoU concepts have recently begun to appear in learning-based models as well, including probabilistic regression~\cite{nattrask_pu} and PoU-augmented DeepONets~\cite{goswami2024learning}. These developments show that the trunk network’s representation is not just a side effect of training, it in fact directly affects how well the operator is learned. 

Building on these developments, in this work, we propose a more general way to bring PoU  structure into physics-informed operator learning. In particular, we propose the \underline{P}hysics-\underline{i}nformed \underline{P}artition \underline{P}enalty Deep Operator \underline{Net}work (PIP$^{2}$ Net). PIP$^{2}$ Net retains the branch--trunk architecture of DeepONet and the physics-based loss used in PI-DeepONet, and introduces a \underline{partition penalty} ($P^{2}$) applied to the trunk outputs. This penalty encourages the trunk outputs to satisfy a global normalization condition, either on their values or on their magnitudes. Intuitively, this makes the collection of trunk outputs behave more like a coordinated family of basis functions, similar in spirit to PoU used in classic numerical method.
This allows us to separate the effects of physics-based losses and partition-based regularization, and to see how they interact in the full PIP$^{2}$ Net. We then study these models on three representative nonlinear PDEs: the viscous Burgers equation, the Allen–Cahn equation, and a diffusion–reaction system. 

The rest of the paper is organized as follows. In Section \ref{sec:pip-net}, we first review DeepONet and PI-DeepONet and then introduce the partition penalty and the resulting PIP$^{2}$ Net architecture. We then show numerical results for the Burgers, Allen–Cahn, and diffusion–reaction equations, comparing DeepONet, PI-DeepONet, POU-DeepONet, and PIP$^{2}$ Net in Section \ref{sec:nu}. Finally, in Section \ref{sec:conclusion}, we summarize the findings and discuss possible extensions in the future work.

\section{Physics-informed Partition Penalty Deep Operator Network (PIP$^2$ Net) \label{sec:pip-net}}
\subsection{Phisics-informed Deep Operator Network (PI-DeepONet)}
Lu et al. \cite{lu2021learning} proposed the general deep operator network (DeepONet) that relies on the universal operator approximation theorem \cite{chen1995universal} that describes the learnability of nonlinear operator mappings between infinite-dimensional function spaces.
To illustrate the DeepONet framework for a general partial differential equation (PDE), consider a PDE of the form
\begin{equation}
\mathcal{F}\big(u(x,t);\kappa\big) = 0, \qquad (x,t)\in \Omega\times[0,T], \label{eqb:pde}
\end{equation}
where $\mathcal{F}$ denotes a (possibly nonlinear) differential operator, and 
$\kappa$ represents the input information associated with the problem such as initial or boundary data, forcing terms, or physical parameters.
DeepONet aims to learn the corresponding solution operator
\begin{align}
\mathcal{G}(\kappa) = u(\cdot;\kappa),    
\end{align}
which maps the input $\kappa$to the full solution field $u$.
A DeepONet can be interpreted as a class of neural network models to approximate $\mathcal{G}$.
In particular, we choose the notation $G_\theta$ to represent the DeepONet approximation of $\mathcal{G}$, 
\begin{align}
    \mathcal{G} \approx G_\theta,
\end{align}
where $\theta$ is the set of trainable network parameters. Then a DeepONet usually consists of two subnetworks \cite{lu2021learning,lu2021deepxde}: the branck net and the trunck net, where the estimator at $\mathbf{x} \in M$ is obtained as the inner product of their outputs:
\begin{align}
G_\theta(\kappa(\Xi))(\mathbf{x}) = \sum_{k=1}^p br_k(\kappa(\xi_1), \kappa(\xi_2), \ldots, \kappa(\xi_m)) \, tr_k(\mathbf{x}) + br_0,    
\end{align}
here $br_0 \in \mathbb{R}$ is a bias, $\{br_1, br_2, \ldots, br_p\}$ are the $p$ outputs of the branch net, and $\{tr_1, tr_2, \ldots, tr_p\}$ are the $p$ outputs of the trunk net. 
More specifically, the branch network represents $\kappa$ in a discrete format: 
\begin{align}
\kappa(\Xi) = \{\kappa(\xi_1), \kappa(\xi_2), \ldots, \kappa(\xi_m)\}, 
\end{align}
where $\kappa(\Xi)$ is a vector which consist the evaluation of input functions at different sensors $\Xi = \{\xi_1, \xi_2, \ldots, \xi_m\}$.
The trunk network  may take as input a spatial location $\mathbf{x} \in M$\cite{lu2021learning,lu2021deepxde,wang2021learning} which yields the following,
\begin{equation}
\operatorname{tr}({\bf x})
   = \bigl(\operatorname{tr}_{1}({\bf x}),\dots,
           \operatorname{tr}_{p}({\bf x})\bigr)\in\mathbb{R}^{p},
           \label{eq:trunk-net}
\end{equation}
where ${\bf x}$ denotes the spatial–temporal coordinate.  
Our goal is to predict the solution corresponding to a given $\kappa$, which is also evaluated at the same input location $\mathbf{x} \in M$. 
The representation of $\kappa$ through pointwise evaluations at arbitrary sensor locations provides additional flexibility, both during training and in prediction, particularly when $\kappa$ is available only through its values at these sensor points.

Given the training data with labels in the two sets of inputs and outputs:
\begin{align}
&  \text{input:}  &\Big\{\kappa^{(k)}, \Xi, (\mathbf{x}_i^{(k)})_{i=1,\ldots,N}\Big\}_{k=1,\ldots,N_{\text{data}}}
    \\
 &   \text{output:}
&\{u(\mathbf{x}_i^{(k)}; \kappa^{(k)})\}_{i=1,\ldots,N,\; k=1,\ldots,N_{\text{data}}}.
\end{align}
we minimize a loss function that measures the discrepancy between the true solution operator $G(\kappa^{(k)})$ and its DeepONet approximation $G_\theta(\kappa^{(k)})$:  
\begin{align}
\widehat{\mathcal{L}}_{\text{data}}(\theta) 
= \frac{1}{N_{\text{data}}N} \sum_{k=1}^{N_{\text{data}}} \sum_{i=1}^N 
\Big| G_\theta(\kappa^{(k)}(\Xi))(\mathbf{x}_i^{(k)}) 
     - G(\kappa^{(k)})(\mathbf{x}_i^{(k)}) \Big|^2,
\end{align}
where $X^{(k)} = \{\mathbf{x}_1^{(k)}, \ldots, \mathbf{x}_N^{(k)}\}$ denotes the set of $N$ evaluation points in the domain of $G(\kappa^{(k)})$.  
In practice, however, the analytical solution of the PDE is not available. Instead, we rely on numerical simulation obtained from high fidelity numerical solvers which can be denoted as: 
\begin{align}
\big\{\hat{u}(\mathbf{x}_i^{(k)}; \kappa^{(k)}) \big\}_{i=1,\ldots,N,\; k=1,\ldots,N_{\text{data}}}.    
\end{align}
Then the practical training tends to minimize the loss function  
\begin{align}
{\mathcal{L}}_{\text{data}}(\theta) 
= \frac{1}{N_{\text{data}}N} \sum_{k=1}^{N_{\text{data}}} \sum_{i=1}^N 
\Big| G_\theta(\kappa^{(k)}(\Xi))(\mathbf{x}_i^{(k)}) 
     - \hat{u}(\mathbf{x}_i^{(k)}; \kappa^{(k)}) \Big|^2.
\end{align}

The Physics-Informed Deep Operator Network (PI-DeepONet) \cite{wang2021learning} integrates the idea of physics-informed neural networks (PINN) \cite{raissi2019physics} with the DeepONet framework. 
The PI-DeepONet introduces the additional PDE residual loss function and boundary condition loss function.
Consequently, the loss function for PI-DeepONet is defined as
\begin{align}
\mathcal{L}(\theta) 
= w_{\text{data}} \mathcal{L}_{\text{data}}(\theta) 
+ w_{physics} \mathcal{L}_{physics}(\theta) 
+ w_{\text{bc}} \mathcal{L}_{\text{bc}}(\theta)
\label{loss-pi-don}
\end{align}

where $\mathcal{L}_{\text{physics}}(\theta)$ enforces the PDE residual,
$\mathcal{L}_{\text{bc}}(\theta)$ enforces the boundary conditions, and $\mathcal{L}_{\text{data}}(\theta)$ accounts for available observational data, which also includes the initial condition. The weights $w_{\text{data}}, w_{\text{physics}}, w_{\text{bc}}$ are tunable hyperparameters that balance the contributions of the individual loss terms.
The boundary-condition loss yields the following: 
\begin{align}
\mathcal{L}_{BC}(\theta)
&= \frac{1}{NP}\sum_{i=1}^{N}\sum_{j=1}^{P}
\left|
G_\theta(u^{(i)})(0,t^{(i)}_{bc,j})
- G_\theta(u^{(i)})(1,t^{(i)}_{bc,j})
\right|^{2} \nonumber \\
&\quad
+ \frac{1}{NP}\sum_{i=1}^{N}\sum_{j=1}^{P}
\left|
\partial_x G_\theta(u^{(i)})(0,t^{(i)}_{bc,j})
- \partial_x G_\theta(u^{(i)})(1,t^{(i)}_{bc,j})
\right|^{2}
\end{align}
which enforces periodicity and consistency of spatial derivatives, while the PDE 
residual term
\begin{align}
\mathcal{L}_{physics}(\theta)
&= \frac{1}{NR}\sum_{i=1}^{N}\sum_{j=1}^{R}
\Big|
\mathcal{F}\big(
G_\theta(u^{(i)});\, 
x^{(i)}_{r,j}, t^{(i)}_{r,j}
\big)
\Big|^{2},
\end{align}
guarantees that the predicted solution satisfies the PDEs \eqref{eqb:pde} at a set of residual points.

\subsection{The {Partition Penalty} ($P^{2}$) in Trunk Net}
The partition of unity (PoU) framework has a long and influential history in scientific computing~\cite{pou, rbf_pu_pde, rbf_pu_basket, rbf_pu_precond, rbf_pu_convection, SWJCP2018}, where it provides a principled mechanism for blending local approximations into globally smooth representations. Its flexibility, numerical stability, and compatibility with meshfree discretizations have also motivated a growing number of applications in scientific machine learning. Notably,~\cite{nattrask_pu} adopts PoU ideas for probabilistic regression, while~\cite{goswami2024learning} incorporates PoU constructions within the Deep Operator Network (DeepONet) architecture to enhance expressivity and local adaptability.

Motivated by these developments, we introduce the \emph{P}artition \emph{P}enalty ($P^{2}$), a mechanism that extends the PoU philosophy to physics informed operator learning framework by explicitly regularizing the trunk network. Conceptually, $P^{2}$ encourages the trunk outputs to behave like a coordinated collection of basis functions, analogous to PoU weight functions. Practically, this is achieved by augmenting the loss function~\eqref{loss-pi-don} with an additional penalty term applied to the trunk network~\eqref{eq:trunk-net}. The goal is to promote a global structure among the trunk outputs that supports stable and interpretable operator approximation.
Formally, the {partition penalty} enforces a normalization constraint of the form
\begin{equation}
\sum_{j=1}^{p} \operatorname{tr}_{j}({\bf x})
   = c(\theta),
\end{equation}
where the normalization factor $c(\theta)$ may be prescribed {a priori} or depend on trainable parameters~$\theta$. A widely used and practically effective choice is the unit normalization
\begin{equation}
c(\theta) \equiv 1 ,
\end{equation}
which forces the collection $\{\operatorname{tr}_{j}\}_{j=1}^{p}$ to collectively form a partition over the evaluation domain.
In situations where sign changes in the trunk outputs are desirable, for example, when representing mixed-mode or oscillatory structures, a natural surrogate of the penalty enforces normalization on their magnitudes yields the following:
\begin{equation}
\sum_{j=1}^{p} \bigl\| \operatorname{tr}_{j}({\bf x}) \bigr\|
   = c(\theta),
\end{equation}
where $\|\cdot\|$ denotes the norm of interest. This preserves the essential balance of the normalization while allowing each trunk component to take positive or negative values.

The introduction of $P^{2}$ introduces several important advantages. First, by enforcing a global normalization, it mitigates pathological behaviors in which a subset of trunk outputs dominates while others collapse toward zero; such imbalance is a known issue in operator networks, especially for operators with multiscale or spatially heterogeneous structure. Second, a near-partition constraint improves the conditioning of the operator map by ensuring that all trunk outputs contribute meaningfully and comparably during training. Finally, the $P^{2}$ mechanism encourages the trunk network to generate basis-like functions whose collective behavior resembles classical PoU systems.

\subsection{The Physics-Informed Partition-Penalty Deep Operator Network (PIP\texorpdfstring{$^{2}$}{2} Net)}
\begin{figure}
    \centering
    \includegraphics[width=\linewidth]{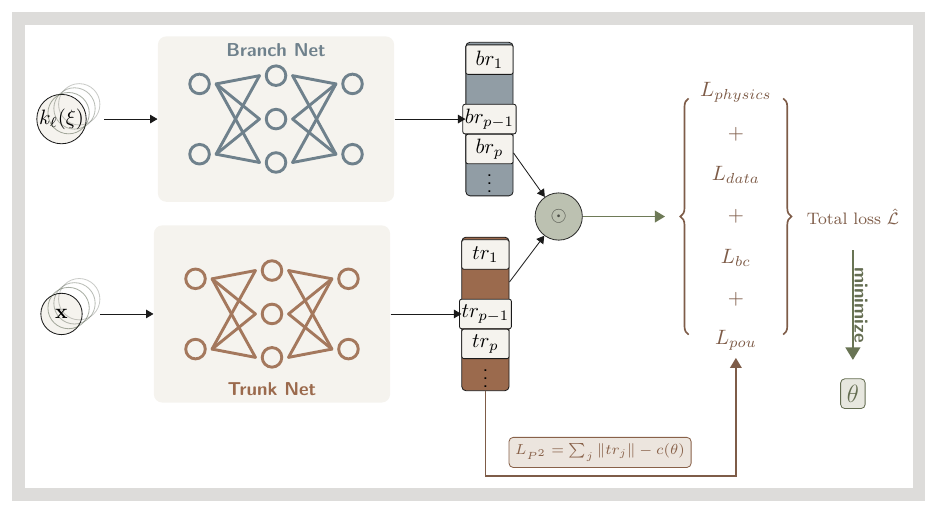}
    \caption{Illustration of PIP$^2$ Net}
    \label{fig:pip2-net}
\end{figure}

Building on the PI-DeepONet framework and the partition-penalty ($P^{2}$) regularization of the trunk network, 
we propose the \emph{P}hysics-\emph{i}nformed \emph{P}artition-\emph{P}enalty Deep Operator Network (PIP$^{2}$ Net). 
This architecture combines physics-informed operator learning framework with PoU inspired by  structural regularization. Figure \ref{fig:pip2-net} illustrates the PIP$^2$ Net framework.
In general, the PIP$^{2}$ Net retains the standard DeepONet representation
\begin{equation}
G_\theta(\kappa(\Xi))(\mathbf{x})
= \sum_{k=1}^{p} br_k(\kappa(\Xi))\, tr_k(\mathbf{x}) + br_0,
\end{equation}
where the branch network encodes the input function $\kappa$ and the trunk network provides a set of basis-like outputs $\{tr_k(\mathbf{x})\}_{k=1}^p$.  
The physics-informed component imposes the PDE residual, boundary conditions, and available data through the loss~\eqref{loss-pi-don}, ensuring that the predicted operator adheres to the governing equations~\eqref{eqb:pde}.

To further improve the stability and representational structure of the trunk outputs, 
PIP$^{2}$ incorporates the partition penalty ($P^{2}$) introduced in the previous subsection.
Specifically, at each spatial--temporal coordinate $\mathbf{x}$, the trunk outputs are constrained by the P$^2$ loss:
\begin{equation}
\mathcal{L}_{P^2} = \sum_{j=1}^{p} \| tr_{j}(\mathbf{x}) \| - c(\theta),
\end{equation}
with $c(\theta)$ either a prescribed constant or a learnable parameter.  
This penalty loss includes the PoU-like structure among the trunk outputs, preventing mode collapse and reducing ill-conditioning in the learned operator representation. Then the combined loss function for PIP$^{2}$ Net therefore takes the form
\begin{equation}
\mathcal{L}_{\text{PIP}^{2}}(\theta)
= w_{\text{data}}\mathcal{L}_{\text{data}}(\theta)
+ w_{physics}\mathcal{L}_{{physics}}(\theta)
+ w_{\text{bc}}\mathcal{L}_{\text{bc}}(\theta)
+ \lambda_{P^{2}} \mathcal{L}_{P^{2}}(\theta),
\end{equation}
where $w_{\text{data}}, w_{physics}, w_{\text{bc}}$ are pre-determined or adaptive coefficients and $\lambda_{P^{2}}$ is the regularization coefficient. 
By integrating physics-informed losses with partition-based trunk regularization, the PIP$^{2}$ Net provides a consistent operator-learning framework that enforces the governing PDE and boundary conditions with specific stabilization of the trunk representation. 

\section{Numerical Tests \label{sec:nu}}
To evaluate the proposed PIP$^{2}$ Net, we test a series of numerical examples for three representative nonlinear PDEs: the Burgers equation, the Allen–Cahn equation, and a diffusion–-reaction equation. As summarized in Table~\ref{tab:models_comparison}, our comparisons include four different operator learning frameworks: DeepONet, PI-DeepONet, POU-DeepONet, and the proposed PIP$^{2}$ Net.
In all test problems, we provide quantitative comparisons of different models using the standard relative $L^{2}$ error between the predicted solution $\widehat{u}$ and the benchmark solution $u$:
\begin{align}
\mathcal{E}_{L^2}
=
\frac{\displaystyle
\left( \int_0^T  \int_\Omega
|u(x,t) - \widehat{u}(x,t)|^{2} dx dt
\right)^{1/2}
}{
\displaystyle
\left( \int_0^T \int_\Omega
|u(x,t)|^{2}  dx  dt
\right)^{1/2}
},
\end{align}
where $\Omega$ denotes the spatial domain and $[0,T]$ is the time interval over which the solution is evaluated.

\begin{table*}[htp!]
\centering
\caption{Summary of models used in numerical tests.}
\label{tab:models_comparison}
\renewcommand{\arraystretch}{1.2}
\begin{tabularx}{\textwidth}{lXccc} 
\hline
\textbf{Acronym} & \textbf{Full Name} & \textbf{Physics-informed} & \textbf{Partition} & \textbf{Normlization}\\
\hline
DeepONet     & \uline{Deep} \uline{O}perator \uline{Net}work  & N/A   & N/A    & N/A    \\
PI-DeepONet       & \uline{P}hysics-\uline{i}nformed  \uline{Deep} \uline{O}perator \uline{Net}work   & Yes  & N/A    & N/A  \\
POU-DeepONet      & \uline{P}artition \uline{O}f  \uline{U}nity  \uline{Deep} \uline{O}perator \uline{Net}work     & N/A   &  Yes  & N/A  \\
PIP$^2$ Net      &\uline{P}hysics-\uline{i}nformed \uline{P}artition  \uline{P}enalty  {Deep} {O}perator \uline{Net}work     & Yes  & Yes   & Yes \\
\hline
\end{tabularx}
\end{table*}

\subsection{One-dimensional Burgers' Equation \label{sec:burgers}}
We first consider the one-dimensional viscous Burgers’ equation which is a standard model for nonlinear advection and shock dynamics \cite{li2021fourier,lu2021deepxde,karniadakis2021physics,lu2025mopinnenkf}. Let $u(x,t)$ denote the velocity field that evolves from an initial profile $u_0(x)$ with viscosity $\nu$. The dynamics are governed by the following PDE with initial condition (IC) and  periodic boundary conditions (BCs):
\begin{subequations}\label{eq:burgers-system}
\begin{align}
    \frac{\partial u}{\partial t}
    + u\,\frac{\partial u}{\partial x}
    &= \nu^{2}\,\frac{\partial^{2} u}{\partial x^{2}},
    \qquad (x,t)\in[-\pi,\pi]\times[0,T],
    \label{eq:burgers-system:a} \\[8pt]
    \text{IC:}\qquad 
    u(x,0) 
    &= u_0(x),
    \qquad x\in[-\pi,\pi],
    \label{eq:burgers-system:b} \\[8pt]
    \text{BCs:}\qquad 
    u(-\pi,t) 
    &= u(\pi,t), \\[-2pt]
    \qquad\qquad 
    \frac{\partial u}{\partial x}(-\pi,t) 
    &= \frac{\partial u}{\partial x}(\pi,t),
    \qquad t\in[0,T].
    \label{eq:burgers-system:c}
\end{align}
\end{subequations}

\begin{figure}[H]
    \centering
    \includegraphics[width=0.5\textwidth]{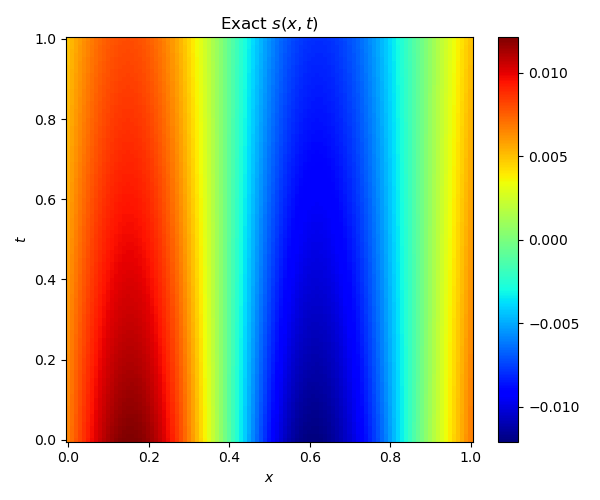}

    \caption{
    Benchmark solution of Burgers' equation with one initial condition.
    }
    \label{fig:bg_exact}
\end{figure}


Operator learning seeks to approximate the solution operator that maps an initial condition $u_0$ to the corresponding solution $u(x,t)$ of \eqref{eq:burgers-system}. In the numerical experiments, we fix the viscosity parameter to $\nu = 0.01$. Following \cite{li2021fourier,lu2021deepxde}, the initial fields $u_0(x)$ are sampled from a Gaussian random field (GRF) with zero mean and a Mat\'ern-type covariance operator,
\begin{equation}
\label{eq:grf}
u_0 \sim 
\mathcal{N}\!\left(0,\; \sigma^{2}\,(-\Delta + \tau^{2} I)^{-\gamma}\right),
\end{equation}
which provides a rich family of spatially correlated and potentially rough initial profiles for training and testing the operator-learning model.




\paragraph{Training and testing data} 
For each initial function $u^{(i)}(x)$, collocation points are sampled as follows. 
At the spatial grid points $x_{ic,j}^{(i)} = x_j$, we randomly sample temporal points to form the boundary sets 
$\{(0,t_{ic,j}^{(i)})\}_{j=1}^{P}$ and $\{(1,t_{ic,j}^{(i)})\}_{j=1}^{P}$, 
as well as the interior residual set 
$\{(x_{r,j}^{(i)},t_{r,j}^{(i)})\}_{j=1}^{Q}$, 
where $P=100$ and $Q=2500$.
A total of $1000$ initial functions are randomly generated from a Gaussian random field 
$\mathcal{N}(0,\,5^{2}(-\Delta+5^{2}I)^{-4})$. 
For each realization, training and testing data are obtained by numerically solving the governing equation with periodic boundary conditions and initial condition $s(x,0)=u(x)$ for $x\in[0,1]$. 
The numerical solutions are computed using the Chebfun package \cite{driscoll2014chebfun} in \textsc{Matlab}, which employs a Fourier spectral discretization in space together with a fourth-order exponential time-differencing scheme for temporal integration.
The Adam optimizer is used to minimize the loss function during training. 
From the $1000$ generated initial functions, $200$ are selected for training.

\paragraph{Neural network architectures} 
Both the branch and trunk networks are implemented as seven-layer fully connected neural networks, with 100 neurons in each hidden layer and the hyperbolic tangent ($\tanh$) activation function. The branch network takes the discretized initial condition as input, while the trunk network takes the spatial coordinates as input.


\paragraph{Neural network parameters} 

We perform a grid search over $w_{\text{data}} \in \{1,5,10,20,50,100\}$ and $\lambda_{p^2} \in \{0.25,0.5,0.75,1\}$ by minimizing the loss function. The optimal values are found to be $w_{\text{data}}=20$ and $\lambda_{p^2}=0.5$. Following the settings in \cite{wang2021learning}, the weights for the physics loss and boundary-condition loss are set to $w_{\text{physics}}=1$ and $w_{\text{bc}}=1$, respectively. The model is trained using a learning rate of $10^{-3}$ for $10{,}000$ iterations.

\begin{table}[H]
\centering
\caption{Comparison of the relative $L^2$ error $\mathcal{E}_{L^2}$ for different operator-learning models for the one-dimensional Burgers’ equation.\label{tab:burgers-l2} 
}
\label{table:loss-comparison}
\renewcommand{\arraystretch}{1.3}
\begin{tabular}{lcccc}
\hline
\textbf{Model} 
& \textbf{DeepONet} 
& \textbf{POU-DeepONet} 
& \textbf{PI-DeepONet} 
& \textbf{PIP$^{2}$ Net} \\
\hline
$\mathcal{E}_{L^2}$ 
& 2.44$\times$10$^{-1}$ 
& 2.05$\times$10$^{-1}$ 
& 1.01$\times$10$^{-1}$ 
& 9.94$\times$10$^{-2}$ \\
\hline
\end{tabular}
\end{table}

Table~\ref{tab:burgers-l2} compares the relative $L^2$ error $\mathcal{E}_{L^2}$ of different operator-learning models for the one-dimensional Burgers’ equation. The baseline DeepONet shows the largest error, indicating limited capability in accurately capturing nonlinear spatiotemporal dynamics using data alone. Introducing a partition-of-unity structure in POU-DeepONet yields a moderate reduction in error, suggesting that localized representations improve approximation quality but remain insufficient without additional physical guidance. Enforcing physics constraints through PI-DeepONet leads to a substantial improvement, reducing the relative error by approximately a factor of two compared to POU-DeepONet. The proposed PIP$^{2}$ Net achieves the smallest $L^2$ error among all models which shows that the combination of physics-informed learning and partition-penalty regularization further improves the results. 

\begin{table}[H]
\centering
\caption{Pointwise absolute errors at selected spatial locations $x$ at time $t = 1$ for different operator-learning models on the one-dimensional Burgers’ equation. \label{tab:burgers-pointwise} }
\renewcommand{\arraystretch}{1.25}
\setlength{\tabcolsep}{4pt}  
\begin{tabular}{lccccccc}
\hline
\textbf{Model} 
& $x=0.2$ 
& $x=0.4$ 
& $x=0.6$ 
& $x=0.8$ 
& $x=1$ 
& \textbf{Avg. Error} \\
\hline
DeepONet 
& $2.47\times10^{-3}$ 
& $5.10\times10^{-3}$ 
& $7.61\times10^{-3}$ 
& $9.93\times10^{-3}$ 
& $1.21\times10^{-2}$ 
& $7.40\times10^{-3}$ \\
\hline
POU-DeepONet 
& $1.58\times10^{-3}$ 
& $1.92\times10^{-3}$ 
& $2.02\times10^{-3}$ 
& $1.92\times10^{-3}$ 
& $1.66\times10^{-3}$ 
& $2.01\times10^{-3}$ \\
\hline
PI-DeepONet 
& $1.00\times10^{-4}$ 
& $4.19\times10^{-4}$ 
& $5.24\times10^{-4}$ 
& $6.81\times10^{-4}$ 
& $9.65\times10^{-4}$ 
& $5.74\times10^{-4}$ \\
\hline
PIP$^{2}$ Net 
& $1.86\times10^{-4}$ 
& $3.65\times10^{-4}$ 
& $5.83\times10^{-4}$ 
& $7.67\times10^{-4}$ 
& $8.49\times10^{-4}$ 
& $5.56\times10^{-4}$ \\
\hline
\end{tabular}
\end{table}
Table~\ref{tab:burgers-pointwise} reports pointwise absolute errors at selected spatial locations at the final time $t=1$ that provides a localized comparison of model performance. Consistent with Table~\ref{tab:burgers-l2}, DeepONet exhibits increasing errors toward the right boundary due to accumulated inaccuracies in nonlinear transport regions. POU-DeepONet reduces pointwise errors over the domain and physics-informed models further improve accuracy, with PI-DeepONet achieving uniformly lower errors at all locations. The proposed PIP$^{2}$ Net attains the smallest average error and maintains low pointwise errors, particularly near $x=1$.

\begin{figure}[H]
    \centering
    \includegraphics[width=0.32\textwidth]{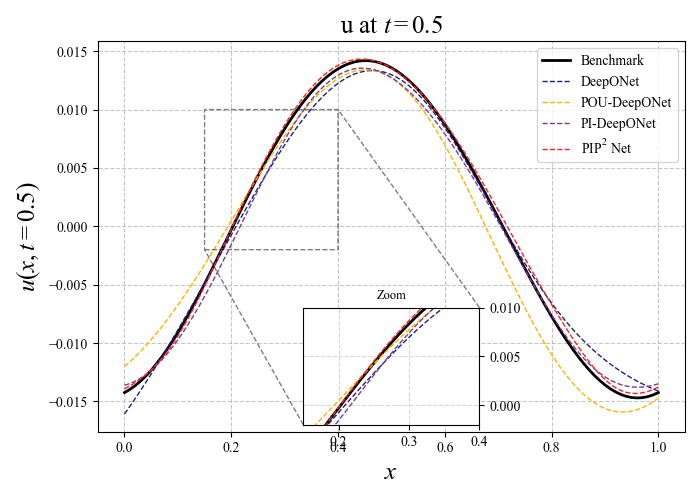}\hfill
    \includegraphics[width=0.32\textwidth]{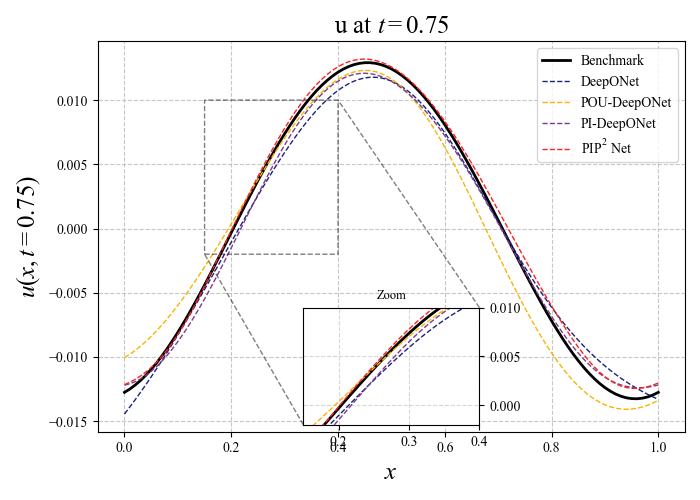}\hfill
    \includegraphics[width=0.32\textwidth]{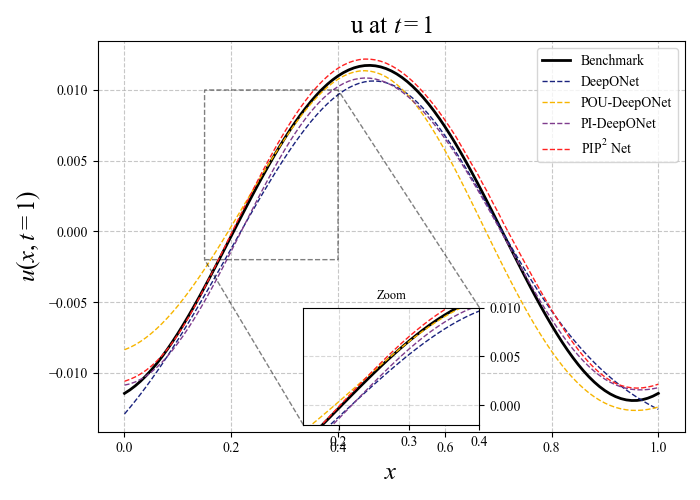}

    \caption{
Comparison of the benchmark and model solutions in Table~\ref{tab:models_comparison} for Burgers’ equation at three different time instants:
(left) $t=0.05$, (middle) $t=0.75$, and (right) $t=1.0$.
    }
    \label{fig:burgers-curve-time}
\end{figure}
As illustrated in Figure~\ref{fig:burgers-curve-time}, all models capture the solution reasonably well at early time, while discrepancies become larger at later times. Among all models, DeepONet exhibits noticeable phase shifts and amplitude errors, particularly near regions of steep gradients. POU-DeepONet improves local accuracy but still shows visible deviations near the right boundary. Physics-informed models achieve closer agreement with the benchmark across all time instants, with PI-DeepONet reducing phase and amplitude errors. The proposed PIP$^{2}$ Net most accurately reproduces the benchmark solution at all three time instances that are consistent with Tables~\ref{tab:burgers-l2} and \ref{tab:burgers-pointwise}.

\begin{figure}[H]
    \centering
    \includegraphics[width=0.32\textwidth]{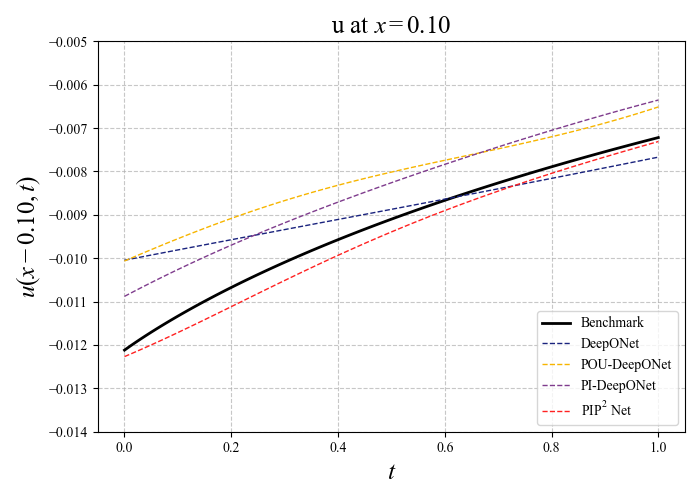}\hfill
    \includegraphics[width=0.32\textwidth]{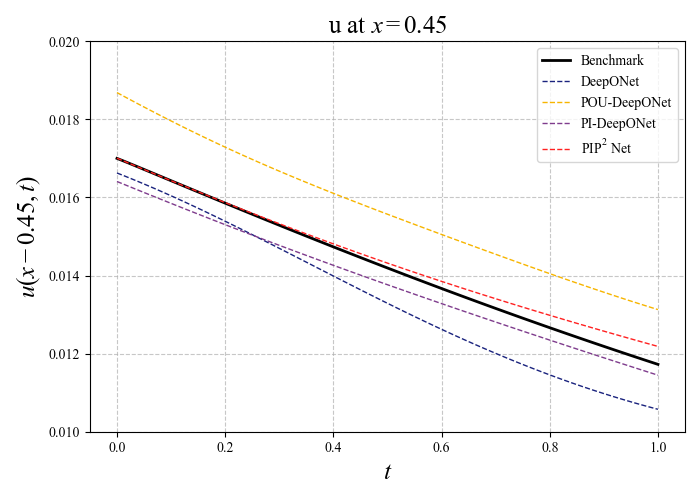}\hfill
    \includegraphics[width=0.32\textwidth]{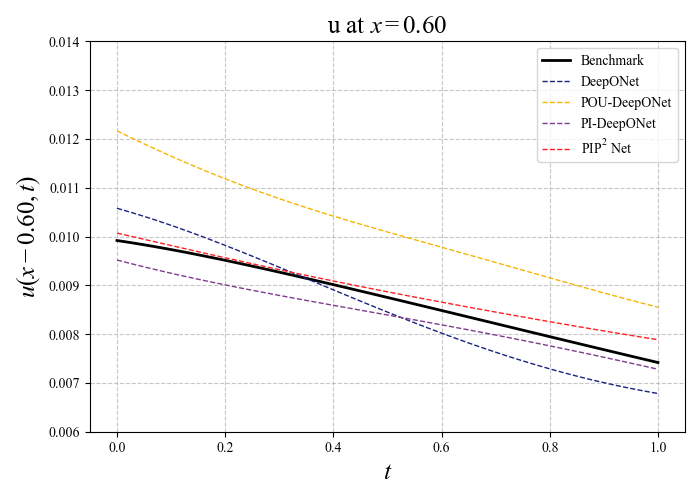}

    \caption{
Comparison of the benchmark and model solutions in Table~\ref{tab:models_comparison} for Burgers’ equation at three different locations:
    $x = 0.1$ (left), $x = 0.45$ (middle), and $x = 0.60$ (right).  
    }
    \label{fig:bg_spatial_slices}
\end{figure}
Consistent with the temporal slices in Figure~\ref{fig:burgers-curve-time}, Figure~\ref{fig:bg_spatial_slices} shows that discrepancies among models become more remarkable at locations with step gradient, i.e., $x=0.45$ and $0.6$. While all models perform similarly at $x=0.1$, DeepONet shows larger temporal error at $x=0.45$ and $x=0.60$, and POU-DeepONet shows moderate improvement. Physics-informed models maintain closer agreement with the benchmark across all locations, with PIP$^{2}$ Net most accurately matches the reference solution, consistent with Tables~\ref{tab:burgers-l2} and \ref{tab:burgers-pointwise}.


\begin{figure}[H]
    \centering

    \begin{subfigure}{\textwidth}
        \centering
        \includegraphics[width=\textwidth]{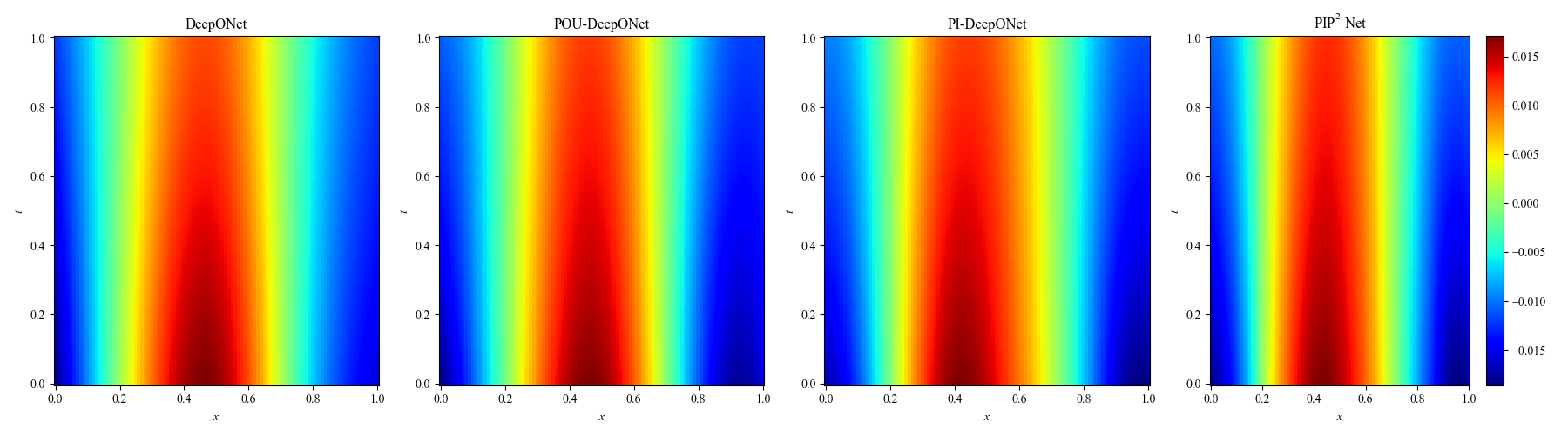}
        \caption{
        Comparison of model solutions in Table~\ref{tab:models_comparison} 
        for the spatiotemporal field of Burgers’ equation.
        }
        \label{fig:bg_prediction}
    \end{subfigure}

    \vspace{1em}

    \begin{subfigure}{\textwidth}
        \centering
        \includegraphics[width=\textwidth]{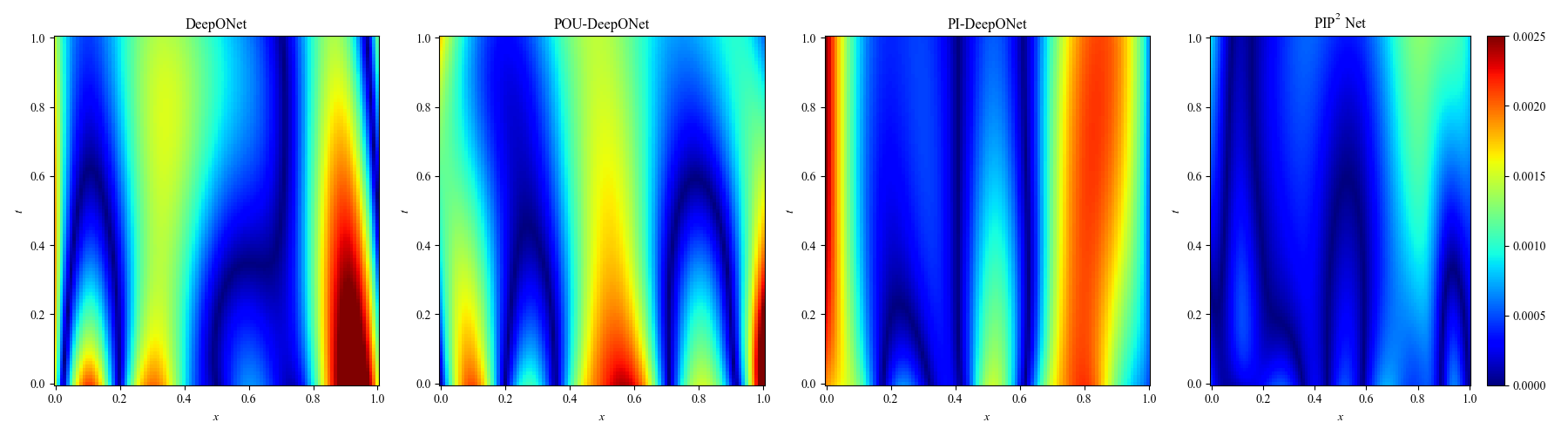}
        \caption{
        Comparison of the model solutions in Table~\ref{tab:models_comparison} 
        for the absolute error of Burgers’ equation.
        }
        \label{fig:bg_error}
    \end{subfigure}

    \caption{
    Predicted spatiotemporal fields and absolute errors for Burgers’ equation 
    obtained from the models summarized in Table~\ref{tab:models_comparison}.
    }
    \label{fig:bg_prediction_error_combined}
\end{figure}
Figure~\ref{fig:bg_prediction_error_combined} compares the predicted spatiotemporal fields and corresponding absolute error. DeepONet and POU-DeepONet show noticeable larger spatiotemporal errors particularly near regions with strong gradients. Physics-informed models substantially reduce these errors. Between the two, the proposed PIP$^{2}$ Net most accurately captures the spatiotemporal evolution and yields the smallest error level that is consistent with  Figures~\ref{fig:burgers-curve-time}--\ref{fig:bg_spatial_slices} and Tables~\ref{tab:burgers-l2} and \ref{tab:burgers-pointwise}.

\subsection{One-dimensional Allen-Cahn Equation  \label{sec:ac}
}
We next consider the one-dimensional Allen--Cahn equation, a prototypical model for phase separation and interface dynamics in materials science. Let $u(x,t)$ denote the order parameter, i.e., the local phase state, with prescribed initial condition $u_0(x)$, governed by the following PDE subject to Dirichlet boundary conditions (BCs):
\begin{subequations}\label{eq:allen-cahn-system}
\begin{align}
\frac{\partial u}{\partial t}
&= \frac{\partial^{2} u}{\partial x^{2}} - \frac{1}{c^{2}},f'(u)
\qquad (x,t)\in[-\pi,\pi]\times[0,1],
\label{eq:allen-cahn-system:a} \\[8pt]
\text{IC:}\qquad
u(x,0)
&= u_0(x),
\qquad x\in[-\pi,\pi],
\label{eq:allen-cahn-system:b} \\[8pt]
\text{BC:}\qquad
u(-\pi,t)
&= u(\pi,t) = 0,
\qquad t\in[0,1].
\label{eq:allen-cahn-system:c}
\end{align}
\end{subequations}
Here $c$ is a model parameter and the nonlinear function $f(u)$ is given by the double-well potential
\begin{equation}
\label{eq:allen-cahn-potential}
f(u) = \frac{(u^{2}-1)^{2}}{4}.
\end{equation}.

\begin{figure}[H]
    \centering
    \includegraphics[width=0.5\textwidth]{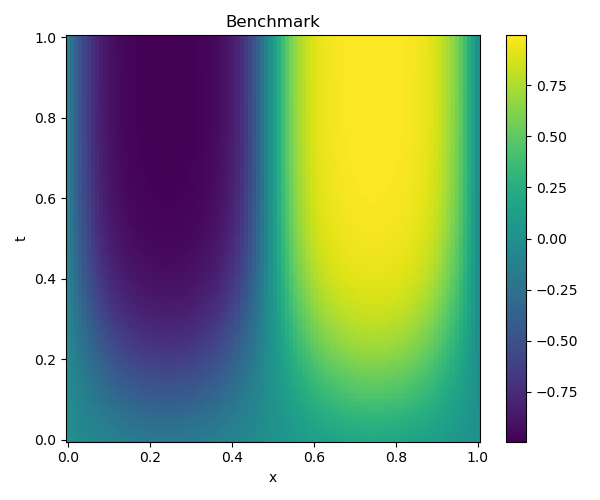}

    \caption{
    Benchmark solution of Allen-Cahn equation with one initial condition.
    }
    \label{fig:ac_exact}
\end{figure}

The Allen--Cahn equation models the evolution of an order parameter during phase transitions by minimizing a free-energy functional composed of gradient and bulk energy terms. Similar to one dimensional Burgers equation in Section \ref{sec:burgers}, operator learning seeks to approximate the solution operator that maps an initial condition $u_0$ (and, when applicable, the parameter $c$) to the corresponding spatiotemporal solution $u(x,t)$ of \eqref{eq:allen-cahn-system}. The benchmark solution of the Allen--Cahn equation is shown in Figure \ref{fig:ac_exact}.

\paragraph{Traning and testing data}
The training and testing datasets are generated by numerically solving the Allen--Cahn equation. The initial condition is prescribed as $u_0(x) = 0.2\sin(x)$, with the parameter $\epsilon^2$ randomly sampled from $[0.1,\,0.5]$. The spatial domain is discretized using 100 uniform grid points, and the temporal interval is resolved with 1000 time steps. An energy-decreasing numerical scheme is employed to obtain the reference solutions. In total, 1000 samples are generated, of which 200 are randomly selected for training.

\paragraph{Neural network architectures}
Similar to the Burgers’ equation test case in Section~\ref{sec:burgers}, both the branch and trunk networks employ seven-layer fully connected architectures with 100 hidden units per layer and hyperbolic tangent (tanh) activations. The branch network takes the discretized initial condition as input, while the trunk network uses the spatial coordinates as input.

\paragraph{Neural network parameters}
A grid search over $w_{\text{data}}\in\{1,5,10,20,50,100\}$ and $w_{p^2}\in\{0.25,0.5,0.75,1\}$ is performed by minimizing the loss function which yields the optimal values $w_{\text{data}}=20$ and $w_{p^2}=0.5$. The model is trained using a learning rate of $10^{-3}$ for $20000$ iterations.



\begin{table}[H]
\centering
\caption{Comparison of the relative $L^2$ error $\mathcal{E}_{L^2}$ for different operator-learning models for the one-dimensional Allen--Cahn equation.\label{tab:ac-l2}}
\renewcommand{\arraystretch}{1.3}
\begin{tabular}{lcccc}
\hline
\textbf{Model} 
& \textbf{DeepONet} 
& \textbf{POU-DeepONet} 
& \textbf{PI-DeepONet} 
& \textbf{PIP$^{2}$ Net} \\
\hline
$\mathcal{E}_{L^2}$ 
& 8.10$\times$10$^{-2}$ 
& 7.75$\times$10$^{-2}$ 
& 2.54$\times$10$^{-2}$ 
& 1.48$\times$10$^{-2}$ \\
\hline
\end{tabular}
\end{table}
Table~\ref{tab:ac-l2} compares the relative $L^2$ error $\mathcal{E}_{L^2}$ of different operator-learning models for the one-dimensional Allen--Cahn equation. Similar to the Burgers’ equation results in Table~\ref{tab:burgers-l2}, the baseline DeepONet exhibits the largest error that shows the limitations of purely data-driven operator learning for nonlinear phase-field dynamics. Introducing a partition-of-unity structure in POU-DeepONet yields a modest improvement, while enforcing physics constraints through PI-DeepONet leads to a substantial reduction in error. The proposed PIP$^{2}$ Net achieves the smallest $L^2$ error among all models that also indicates the combined effect of physics-informed learning and partition-penalty regularization further improves the accuracy. Compared to the Burgers’ equation, the relative improvement of PIP$^{2}$ Net over PI-DeepONet is more remarkable which is due to the fact that the Allen--Cahn dynamics is more sensitive to interface resolution.

\begin{table}[H]
\centering
\caption{Pointwise absolute errors at selected spatial locations $x$ at time $t = 0.9$ for different operator-learning models on the one-dimensional Allen--Cahn equation. \label{tab:ac-pointwise} }
\renewcommand{\arraystretch}{1.25}
\setlength{\tabcolsep}{4pt}  

\begin{tabular}{lccccccl}
\hline
\textbf{Model} 
& $x=0$ 
& $x=0.2$ 
& $x=0.4$ 
& $x=0.6$ 
& $x=0.8$ 
& $x=1$ 
& \textbf{Avg. Error} \\
\hline
DeepONet 
& 5.89$\times$10$^{-3}$ 
& 7.39$\times$10$^{-2}$ 
& 1.32$\times$10$^{-1}$ 
& 1.24$\times$10$^{-1}$ 
& 1.17$\times$10$^{-1}$ 
& 1.06$\times$10$^{-1}$ 
& 1.11$\times$10$^{-1}$ \\
\hline
POU-DeepONet 
& 1.19$\times$10$^{-2}$ 
& 5.23$\times$10$^{-2}$ 
& 7.93$\times$10$^{-2}$ 
& 9.04$\times$10$^{-2}$ 
& 1.01$\times$10$^{-1}$ 
& 1.17$\times$10$^{-1}$ 
& 9.04$\times$10$^{-2}$ \\
\hline
PI-DeepONet 
& 6.94$\times$10$^{-3}$ 
& 1.07$\times$10$^{-2}$ 
& 7.66$\times$10$^{-3}$ 
& 1.99$\times$10$^{-2}$ 
& 2.20$\times$10$^{-2}$ 
& 2.56$\times$10$^{-2}$ 
& 1.85$\times$10$^{-2}$ \\
\hline
PIP$^{2}$ Net 
& 3.68$\times$10$^{-3}$ 
& 8.31$\times$10$^{-3}$ 
& 5.03$\times$10$^{-3}$ 
& 6.35$\times$10$^{-3}$ 
& 6.42$\times$10$^{-3}$ 
& 6.09$\times$10$^{-3}$ 
& 7.17$\times$10$^{-3}$ \\
\hline
\end{tabular}
\end{table}

Table~\ref{tab:ac-pointwise} shows pointwise absolute errors at selected spatial locations at time $t=0.9$, which provides a localized comparison of model performance. Consistent with the Burgers’ equation results in Table~\ref{tab:burgers-pointwise}, DeepONet still shows large pointwise errors for most spatial locations, particularly away from the boundaries. POU-DeepONet reduces these errors but remains less accurate near regions associated with interface motion. Physics-informed models significantly improve pointwise accuracy, with PI-DeepONet achieving uniformly lower errors across the domain. The proposed PIP$^{2}$ Net attains the smallest average error and maintains consistently low pointwise errors at all spatial locations.

\begin{figure}[H]
    \centering
    \includegraphics[width=0.32\textwidth]{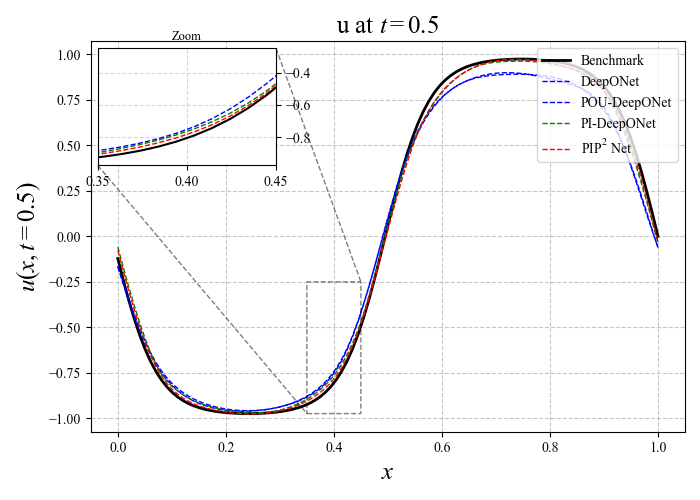}\hfill
    \includegraphics[width=0.32\textwidth]{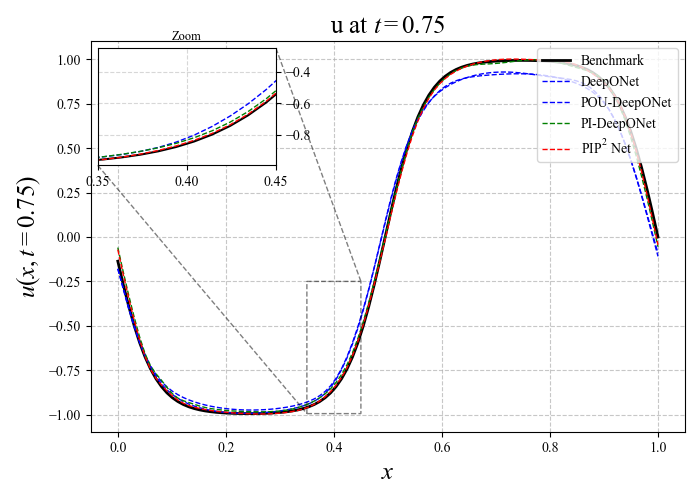}\hfill
    \includegraphics[width=0.32\textwidth]{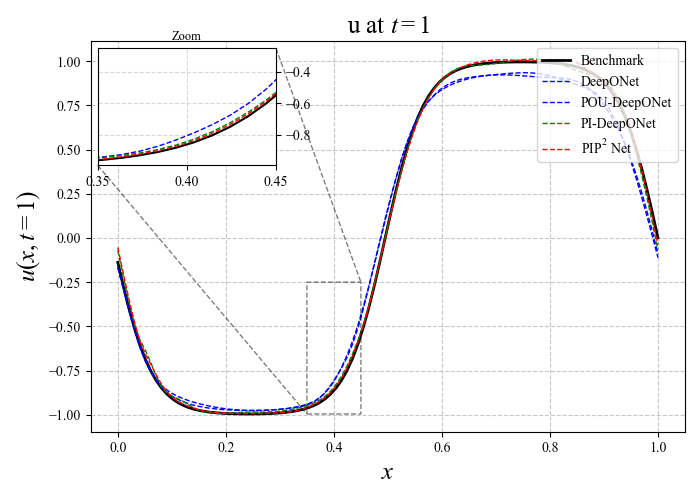}

    \caption{
Comparison of the benchmark and model solutions in Table~\ref{tab:models_comparison} for Allen-cahn equation at three different time instants:
(left) $t=0.05$, (middle) $t=0.75$, and (right) $t=1.0$.
    }
    \label{fig:ac-curve-time}
\end{figure}
As illustrated in Figure~\ref{fig:ac-curve-time}, all models capture the qualitative solution behavior at early time, while discrepancies become more remarkable as the phase separation dynamics evolve. Similar to the temporal results observed for the Burgers’ equation in Figure~\ref{fig:burgers-curve-time}, DeepONet exhibits strong phase lag and amplitude errors at later times. POU-DeepONet improves local accuracy but still deviates from the benchmark solution. Physics-informed models achieve closer agreement across all time instants, with PI-DeepONet reducing temporal errors. The proposed PIP$^{2}$ Net most accurately reproduces the benchmark solution at all three time instances, consistent with the error trends reported in Tables~\ref{tab:ac-l2} and \ref{tab:ac-pointwise}.

\begin{figure}[H]
    \centering
    \includegraphics[width=0.32\textwidth]{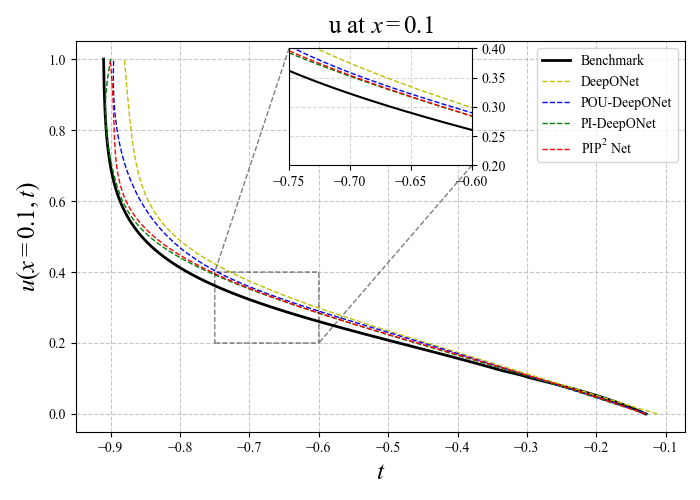}\hfill
    \includegraphics[width=0.32\textwidth]{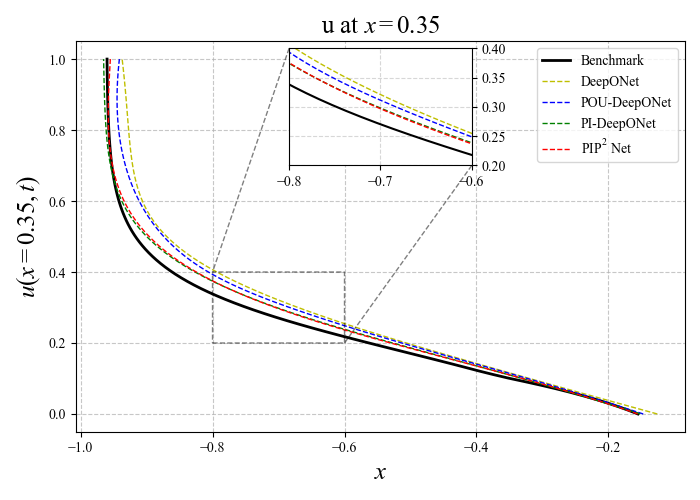}\hfill
    \includegraphics[width=0.32\textwidth]{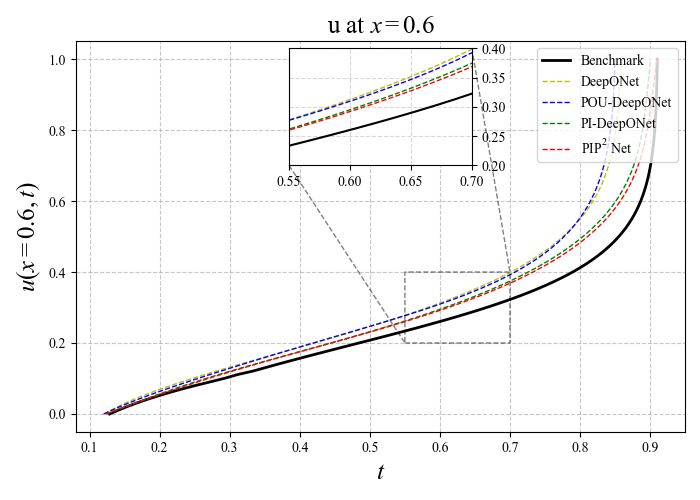}

    \caption{
Comparison of the benchmark and model solutions in Table~\ref{tab:models_comparison} for Allen-cahn equation at three different locations:
    $x = 0.1$ (left), $x = 0.35$ (middle), and $x = 0.60$ (right).  
    }
    \label{fig:AL_spatial_slices}
\end{figure}
Consistent with the temporal behavior, Figure~\ref{fig:AL_spatial_slices} shows that discrepancies among models become more evident at spatial locations influenced by interfacial dynamics. While all models perform comparably at $x=0.1$, larger deviations are observed at $x=0.35$ and $x=0.60$. Similar to the Burgers’ equation case, DeepONet exhibits the largest temporal errors, and POU-DeepONet provides moderate improvement. Physics-informed models maintain closer agreement with the benchmark across all locations, with PIP$^{2}$ Net most accurately matching the reference solution, consistent with the pointwise error statistics in Table~\ref{tab:ac-pointwise}.


\begin{figure}[H]
    \centering

    \begin{subfigure}{\textwidth}
        \centering
        \includegraphics[width=\textwidth]{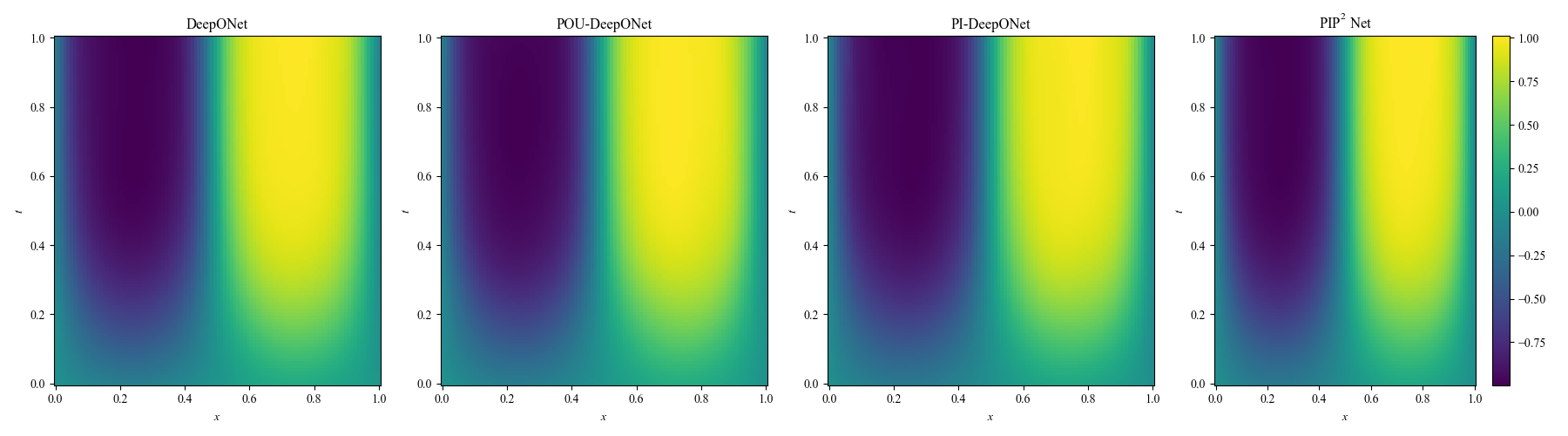}
        \caption{
        Comparison of model solutions in Table~\ref{tab:models_comparison} 
        for the spatiotemporal field of Allen-cahn equation.
        }
        \label{fig:ac_prediction}
    \end{subfigure}

    \vspace{1em}

    \begin{subfigure}{\textwidth}
        \centering
        \includegraphics[width=\textwidth]{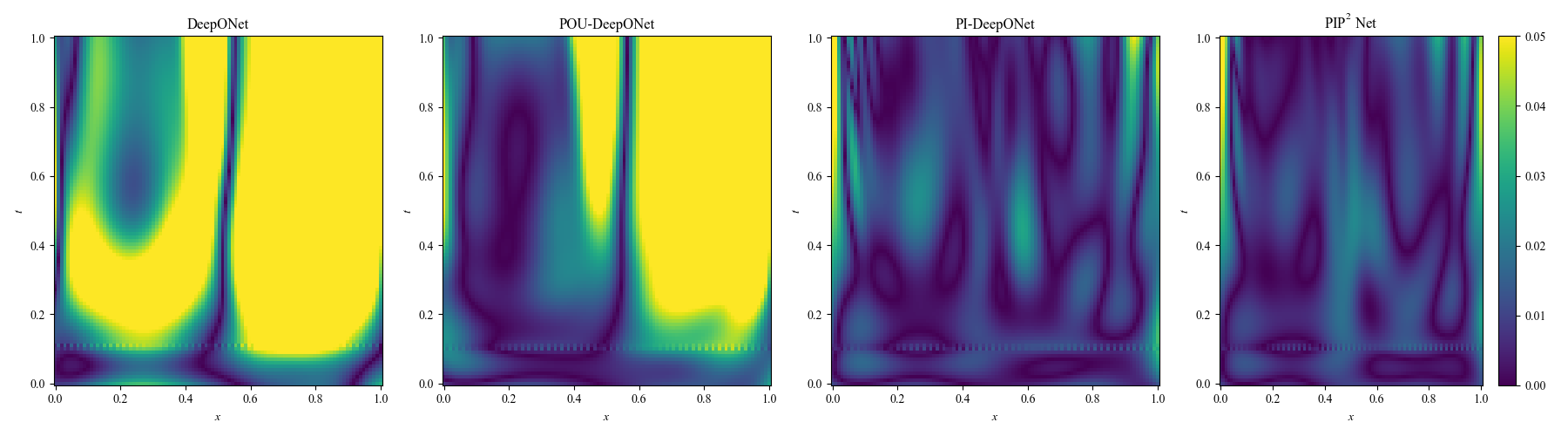}
        \caption{
        Comparison of the model solutions in Table~\ref{tab:models_comparison} 
        for the absolute error of Allen-cahn equation.
        }
        \label{fig:ac_error}
    \end{subfigure}

    \caption{
    Predicted spatiotemporal fields and absolute errors for Allen-cahn equation 
    obtained from the models summarized in Table~\ref{tab:models_comparison}.
    }
    \label{fig:ac_prediction_error_combined}
\end{figure}
Figure~\ref{fig:ac_prediction_error_combined} compares the predicted spatiotemporal fields and corresponding absolute errors for the Allen--Cahn equation. Similar to the spatiotemporal results for the Burgers’ equation in Figure~\ref{fig:bg_prediction_error_combined}, DeepONet and POU-DeepONet exhibit larger spatiotemporal errors, particularly near regions with sharp gradients. Physics-informed models substantially suppress these errors, with PI-DeepONet improving spatiotemporal coherence. Among all methods, the proposed PIP$^{2}$ Net most accurately captures the spatiotemporal evolution and yields the smallest error levels.

\subsection{One-dimensional Diffusion-Reaction Equation \label{sec:dr}}
\begin{figure}[H]
    \centering
    \includegraphics[width=0.5\textwidth]{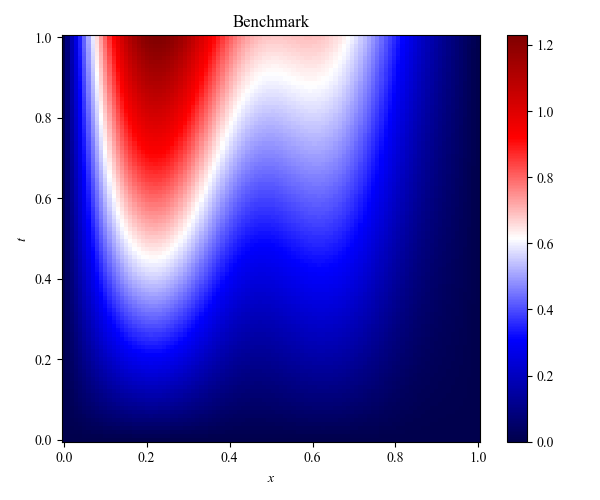}

    \caption{
    Benchmark solution of diffusion-reaction equation with one initial condition.
    }
    \label{fig:dr_exact}
\end{figure}

We finally consider the one-dimensional diffusion--reaction equation, a canonical model for coupled material diffusion and chemical reaction processes \cite{reitz1981study}. Let $s(x,t)$ denote the concentration field with initial condition $s_0(x)$, evolving under diffusion with coefficient $D$, a quadratic reaction term with rate $k$, and an external source term $u(x)$. The dynamics are governed by the following PDE subject to Dirichlet boundary conditions (BCs):
\begin{subequations}\label{eq:diff-react-system}
\begin{align}
    \frac{\partial s}{\partial t}
    &= D\,\frac{\partial^{2} s}{\partial x^{2}} + k\,s^{2} + u(x),
    \qquad (x,t)\in[0,1]\times[0,1],
    \label{eq:diff-react-system:a} \\[8pt]
    \text{IC:}\qquad 
    s(x,0) 
    &= s_0(x),
    \qquad x\in[0,1],
    \label{eq:diff-react-system:b} \\[8pt]
    \text{BC:}\qquad 
    s(0,t) 
    &= s(1,t) = 0,
    \qquad t\in[0,1].
    \label{eq:diff-react-system:c}
\end{align}
\end{subequations}
In this example, homogeneous initial and boundary conditions are imposed, with $s_0(x)=0$ and $s(0,t)=s(1,t)=0$. The diffusion coefficient and reaction rate are fixed at $D=0.01$ and $k=0.01$, respectively. The operator learning models aims to approximate the solution operator that maps a given source term $u(x)$ to the corresponding concentration field $s(x,t)$ satisfying \eqref{eq:diff-react-system}. Following \cite{wang2021learning}, the source terms $u(x)$ are sampled from a Gaussian random field (GRF) with a radial basis function (RBF) covariance kernel,
\begin{equation}
\label{eq:rbf-kernel}
\mathcal{K}(x_1,x_2) = \sigma^{2}\exp\!\left(-\frac{1}{2}\sum_i\left(\frac{x_{1,i}-x_{2,i}}{l_i}\right)^{2}\right),
\end{equation}
which provides a set of smooth source functions for training and testing. The benchmark solution is shown in Figure \ref{fig:dr_exact}.


\paragraph{Training and tesing data}
Consistent with the setups in Sections~\ref{sec:burgers} and \ref{sec:ac}, for each training sample $u^{(i)}$, we uniformly sample $P$ points $\{(x_{u,j}^{(i)}, t_{u,j}^{(i)})\}_{j=1}^{P}$ from the boundary of $[0,1]\times[0,1]$, excluding the final time $t=1$, and $Q$ collocation points $\{(x_{r,j}^{(i)}, t_{r,j}^{(i)})\}_{j=1}^{Q}$, where $x_{r,j}^{(i)}=x_j$ and $\{t_{r,j}^{(i)}\}_{j=1}^{Q}$ are uniformly sampled in $[0,1]$. We set $P=Q=100$ and generate 500 training input functions $u(x)$ by sampling from a Gaussian random field (GRF) with length scale $l=0.2$ using a spatial resolution of $m=100$ equidistant points. The test set consists of 50 independently sampled input functions from the same GRF, with reference solutions computed using a second-order implicit finite difference scheme on a $100\times100$ uniform grid.

\paragraph{Neural network architectures}
Consistent with the architectures used in Sections~\ref{sec:burgers} and \ref{sec:ac}, the branch network takes the input function $u(x)$ and is implemented as a six-layer fully connected neural network with 50 hidden units per layer and hyperbolic tangent (\texttt{tanh}) activations. The trunk network uses the spatial coordinates as input and adopts the same six-layer fully connected architecture with identical activation functions.

\paragraph{Neural network parameters}
Following the same hyperparameter tuning strategy as in Sections~\ref{sec:burgers} and \ref{sec:ac}, a grid search over $w_{p^2}\in\{0.05, 0.1, 0.15, 0.2, 0.5, 1\}$ is performed by minimizing the loss function, yielding the optimal value $w_{p^2}=0.1$. In this example, we use a learning rate of $10^{-3}$, and train the model for $10000$ iterations.

\begin{table}[H]
\centering
\caption{Comparison of the relative $L^2$ error $\mathcal{E}_{L^2}$ for different operator-learning models for the one-dimensional diffusion--reaction equation.\label{tab:dr-l2} }
\renewcommand{\arraystretch}{1.3}
\begin{tabular}{lcccc}
\hline
\textbf{Model} 
& \textbf{DeepONet} 
& \textbf{POU-DeepONet} 
& \textbf{PI-DeepONet} 
& \textbf{PIP$^{2}$ Net} \\
\hline
$\mathcal{E}_{L^2}$ 
& 6.65$\times$10$^{-2}$ 
& 5.10$\times$10$^{-2}$ 
& 4.49$\times$10$^{-2}$ 
& 1.56$\times$10$^{-4}$ \\
\hline
\end{tabular}
\end{table}
Table~\ref{tab:dr-l2} shows the relative $L^2$ error $\mathcal{E}_{L^2}$ for different operator-learning models applied to the one-dimensional diffusion--reaction equation. Similar to the Burgers’ and Allen--Cahn cases, the baseline DeepONet exhibits the largest error that indicates the limited accuracy when learning the solution operator from data alone. Incorporating a partition-of-unity structure in POU-DeepONet leads to a noticeable reduction in error. Physics-informed learning further enhances performance, with PI-DeepONet achieving additional error reduction. Notably, the proposed PIP$^{2}$ Net yields a dramatic improvement, attaining an $L^2$ error several orders of magnitude smaller than the other models.

\begin{table}[H]
\centering
\caption{Pointwise absolute errors at selected spatial locations $x$ at time $t = 1$ for different operator-learning models on the one-dimensional diffusion--reaction equation. \label{tab:dr-pointwise}}
\renewcommand{\arraystretch}{1.25}
\setlength{\tabcolsep}{4pt}  
\begin{tabular}{lccccccc}
\hline
\textbf{Model} 
& $x=0$ 
& $x=0.2$ 
& $x=0.4$ 
& $x=0.6$ 
& $x=0.8$ 
& $x=1$ 
& \textbf{Avg. Error} \\
\hline
DeepONet 
& 1.63$\times$10$^{-2}$ 
& 2.61$\times$10$^{-2}$ 
& 3.59$\times$10$^{-2}$ 
& 4.49$\times$10$^{-1}$ 
& 5.35$\times$10$^{-1}$ 
& 6.08$\times$10$^{-1}$ 
& 4.75$\times$10$^{-1}$ \\
\hline
POU-DeepONet 
& 1.39$\times$10$^{-2}$ 
& 1.11$\times$10$^{-3}$ 
& 6.36$\times$10$^{-3}$ 
& 6.56$\times$10$^{-3}$ 
& 6.71$\times$10$^{-3}$ 
& 1.10$\times$10$^{-2}$ 
& 9.13$\times$10$^{-3}$ \\
\hline
PI-DeepONet 
& 1.37$\times$10$^{-2}$ 
& 1.20$\times$10$^{-3}$ 
& 1.23$\times$10$^{-3}$ 
& 1.22$\times$10$^{-2}$ 
& 1.06$\times$10$^{-2}$ 
& 7.43$\times$10$^{-3}$ 
& 1.36$\times$10$^{-2}$ \\
\hline
PIP$^{2}$ Net 
& 5.82$\times$10$^{-3}$ 
& 7.53$\times$10$^{-3}$ 
& 5.19$\times$10$^{-3}$ 
& 2.46$\times$10$^{-3}$ 
& 1.78$\times$10$^{-3}$ 
& 4.98$\times$10$^{-3}$ 
& 5.55$\times$10$^{-3}$ \\
\hline
\end{tabular}
\end{table}
Table~\ref{tab:dr-pointwise} presents pointwise absolute errors at selected spatial locations at the final time $t=1$, providing a localized comparison of model accuracy. DeepONet exhibits large pointwise errors, particularly toward the right boundary, where the influence of the source term accumulates over time. POU-DeepONet significantly reduces these errors across most spatial locations, while PI-DeepONet further improves accuracy by enforcing the governing PDE residual. Consistent with the average $L^2$ error trends in Table~\ref{tab:dr-l2}, the proposed PIP$^{2}$ Net achieves the smallest average error and maintains uniformly low pointwise errors throughout the domain. Compared to the Burgers’ and Allen--Cahn results, the improvement here is especially large that further indicates the robustness of PIP$^{2}$ Net for source-driven diffusion processes.

\begin{figure}[H]
    \centering
    \includegraphics[width=0.32\textwidth]{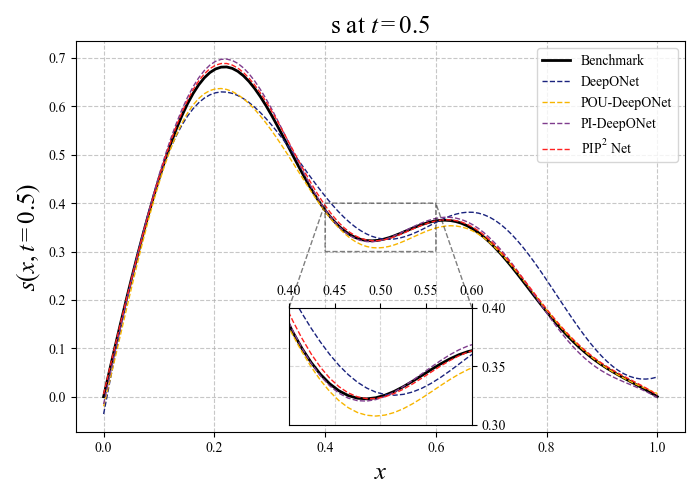}\hfill
    \includegraphics[width=0.32\textwidth]{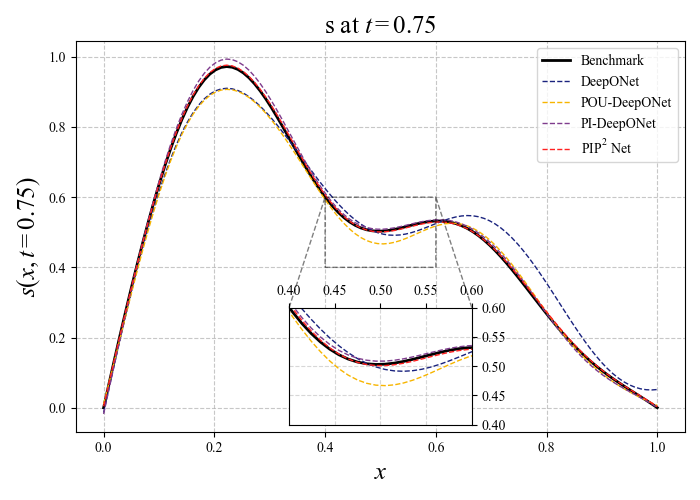}\hfill
    \includegraphics[width=0.32\textwidth]{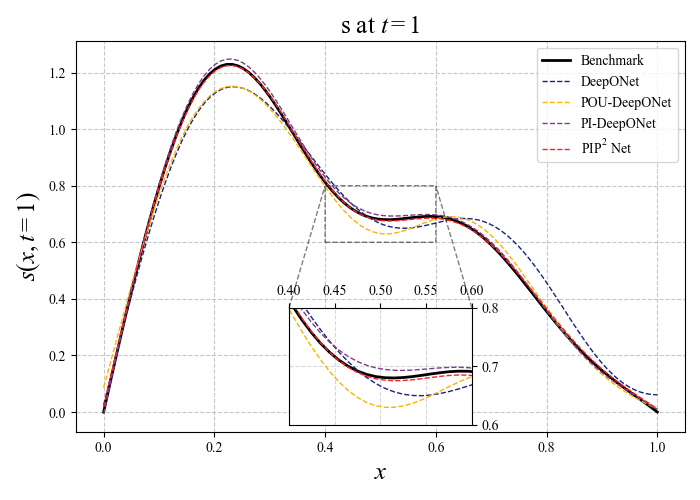}

    \caption{
Comparison of the benchmark and model solutions in Table~\ref{tab:models_comparison} for Diffusion-Reaction equation at three different time instants:
(left) $t=0.50$, (middle) $t=0.75$, and (right) $t=1.0$.
    }
    \label{fig:dr}
\end{figure}
As shown in Figure~\ref{fig:dr}, all models capture the overall solution profile at early time, while discrepancies become more evident as the diffusion--reaction dynamics evolve. DeepONet exhibits noticeable amplitude and phase errors, particularly in regions influenced by the source term, which grow over time. POU-DeepONet improves local accuracy but still deviates from the benchmark near regions of strong curvature. Physics-informed models provide closer agreement across all time instants, with PI-DeepONet reducing both phase and amplitude errors. The proposed PIP$^{2}$ Net most accurately reproduces the benchmark solution at all three time instances, especially in regions highlighted by the zoomed-in views, consistent with the error trends reported in Tables~\ref{tab:dr-l2} and \ref{tab:dr-pointwise} and analogous to the improvements observed for the Burgers’ and Allen--Cahn equation.

\begin{figure}[H]
    \centering
    \includegraphics[width=0.32\textwidth]{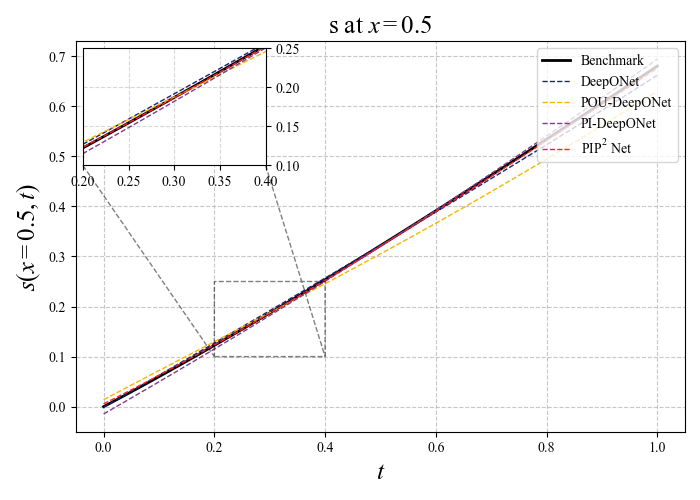}\hfill
    \includegraphics[width=0.32\textwidth]{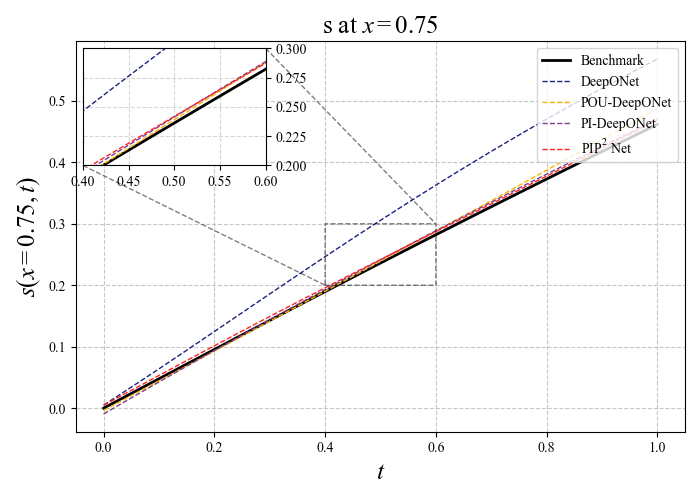}\hfill
    \includegraphics[width=0.32\textwidth]{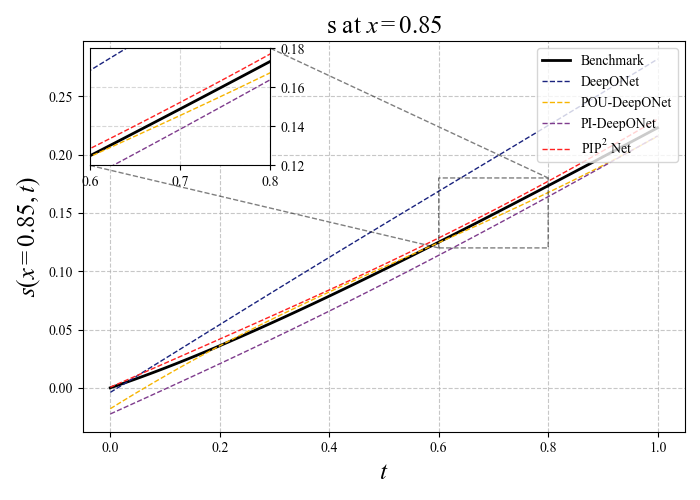}

    \caption{
Comparison of the benchmark and model solutions in Table~\ref{tab:models_comparison} for Diffusion-Reaction equation at three different locations:
    $x = 0.5$ (left), $x = 0.75$ (middle), and $x = 0.85$ (right).  
    }
    \label{fig:dr_spatial_slices}
\end{figure}
Figure~\ref{fig:dr_spatial_slices} presents the temporal evolution of the solution at fixed spatial locations. At $x=0.5$, all models closely follow the benchmark solution. Larger discrepancies emerge at $x=0.75$ and $x=0.85$, where the influence of the source term and reaction effects becomes more pronounced. DeepONet displays the largest temporal deviations, while POU-DeepONet provides moderate improvement. Physics-informed models maintain closer agreement throughout the time interval, with the proposed PIP$^{2}$ Net most accurately tracking the benchmark solution at all locations. This behavior is consistent with the pointwise error trends in Table~\ref{tab:dr-pointwise} and consistent with the results for the Burgers’ and Allen--Cahn equations.


\begin{figure}[H]
    \centering

    \begin{subfigure}{\textwidth}
        \centering
        \includegraphics[width=\textwidth]{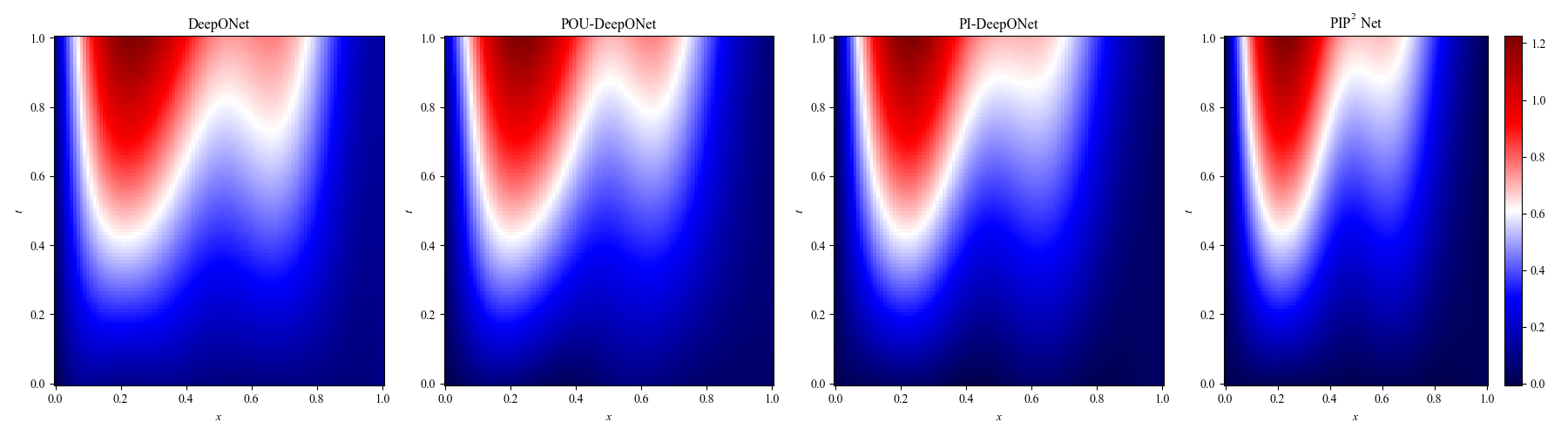}
        \caption{
        Comparison of model solutions in Table~\ref{tab:models_comparison} 
        for the spatiotemporal field of Diffusion-Reaction equation.
        }
        \label{fig:dr_prediction}
    \end{subfigure}

    \vspace{1em}

    \begin{subfigure}{\textwidth}
        \centering
        \includegraphics[width=\textwidth]{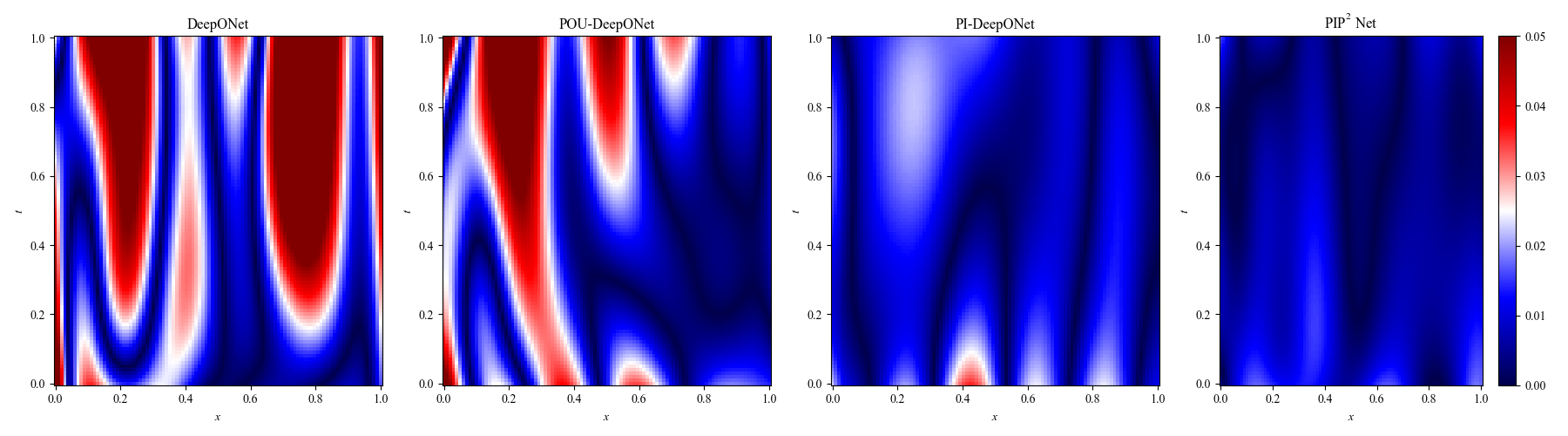}
        \caption{
        Comparison of the model solutions in Table~\ref{tab:models_comparison} 
        for the absolute error of Diffusion-Reaction equation.
        }
        \label{fig:dr_error}
    \end{subfigure}

    \caption{
    Predicted spatiotemporal fields and absolute errors for Diffusion-Reaction equation 
    obtained from the models summarized in Table~\ref{tab:models_comparison}.
    }
    \label{fig:dr_prediction_error_combined}
\end{figure}
Figure~\ref{fig:dr_prediction_error_combined} illustrates the predicted spatiotemporal fields and corresponding absolute error distributions for the diffusion--reaction equation. While all models recover the overall solution structure, DeepONet and POU-DeepONet exhibit substantial spatiotemporal error accumulation, particularly in regions strongly influenced by the source term. Physics-informed models markedly reduces these errors. Between the two, the proposed PIP$^{2}$ Net achieves the most accurate spatiotemporal representation and the lowest error levels. This behavior is consistent with the trends observed in Figures~\ref{fig:dr}–-\ref{fig:dr_spatial_slices} and Tables~\ref{tab:dr-l2} and \ref{tab:dr-pointwise}, and further corroborates the advantages of PIP$^{2}$ Net observed for the Burgers’ and Allen--Cahn equations.

\section{Conclusion and Future Work \label{sec:conclusion}}
In this work, we propose a physics-informed operator-learning framework, termed PIP$^{2}$ Net, which incorporates a Partition Penalty ($P^{2}$) mechanism into the physics-informed DeepONet architecture. The proposed framework tends to improve the structural instability of trunk-network basis functions that can occur in existing operator-learning models and may result in mode imbalance or collapse \cite{kovachki2024data}. The Partition Penalty refers to a partition-of-unity (PoU)–inspired penalty regularization \cite{rbf_pu_precond,lu2021deepxde}, which encourages well-conditioned and coordinated trunk-network outputs, that further enhances the stability  of operator approximation. We evaluate PIP$^{2}$ Net on three representative nonlinear partial differential equations, i.e., the Burgers equation, the Allen–Cahn equation, and the diffusion–reaction equation \cite{evans2022partial,farlow1993partial}. Comparisons with DeepONet, PI-DeepONet, and POU-DeepONet \cite{lu2025mopinnenkf,rbf_pu_basket} show that PIP$^{2}$ Net consistently achieves lower $L^2$ errors for all test cases.

In the future, we plan to investigate several promising research directions. First, we will extend the proposed PIP$^{2}$-Net to substantially more complex physical systems, such as the Navier--Stokes equations (NSEs) \cite{mou2021data,mou2023energy} and the Boussinesq equations \cite{zhang2022convergence,zhang2021fast,zhang2021effects}, where multiscale interactions and chaotic dynamics pose significant challenges for existing operator-learning frameworks. Another important direction is the integration of the $P^{2}$ mechanism into Fourier Neural Operator (FNO) architectures \cite{li2021fourier} to develop PIP$^{2}$ based FNO models with improved spectral stability and improved ability to capture high-frequency dynamics. Additional directions include geometry-aware extensions for problems defined on irregular or parameterized domains, coupling PIP$^{2}$-Net with data assimilation techniques \cite{mou2023combining,chen2020efficient} to enable real-time digital twins \cite{rasheed2020digital,chen2025targeted}, developing probabilistic learning framework \cite{chen2024stochastic,chen2024learning} for uncertainty quantification in sparse observation settings, and deriving theoretical analysis to better understand how the $P^{2}$ regularization improves conditioning, generalization, and operator approximation.



\section*{Data and Code Availability}
The implementation of the proposed method is available at
\url{https://github.com/Hongjin-Mi/PIP2-Net}.
Data supporting the findings of this work can be obtained from the corresponding author upon reasonable request.

\section*{Acknowledgments}
Y.Z is grateful to acknowledge the support of the National Natural Science Foundation of China (NSFC) under Grants No. 12401562 , No. 12571459, and No. 12241103.
\bibliographystyle{amsplain}
\bibliography{references}    

@article{lu2023nsga,
  title={NSGA-PINN: a multi-objective optimization method for physics-informed neural network training},
  author={Lu, Binghang and Moya, Christian and Lin, Guang},
  journal={Algorithms},
  volume={16},
  number={4},
  pages={194},
  year={2023},
  publisher={MDPI}
}

@misc{li2021fourier,
      title={Fourier Neural Operator for Parametric Partial Differential Equations}, 
      author={Zongyi Li and Nikola Kovachki and Kamyar Azizzadenesheli and Burigede Liu and Kaushik Bhattacharya and Andrew Stuart and Anima Anandkumar},
      year={2021},
      eprint={2010.08895},
      archivePrefix={arXiv},
      primaryClass={cs.LG}
}

@article{lu2025evolutionary,
  title={An Evolutionary Multi-objective Optimization for Replica-Exchange-based Physics-informed Operator Learning Network},
  author={Lu, Binghang and Mou, Changhong and Lin, Guang},
  journal={arXiv preprint arXiv:2509.00663},
  year={2025}
}

@article{chen1995universal,
  title={Universal approximation to nonlinear operators by neural networks with arbitrary activation functions and its application to dynamical systems},
  author={Chen, Tianping and Chen, Hong},
  journal={IEEE Transactions on Neural Networks},
  volume={6},
  number={4},
  pages={911--917},
  year={1995},
  publisher={IEEE}
}

@article{lu2021deepxde,
  title={DeepXDE: A deep learning library for solving differential equations},
  author={Lu, Lu and Meng, Xuhui and Mao, Zhiping and Karniadakis, George Em},
  journal={SIAM Review},
  volume={63},
  number={1},
  pages={208--228},
  year={2021},
  publisher={SIAM}
}

@article{raissi2019physics,
  title={Physics-informed neural networks: A deep learning framework for solving forward and inverse problems involving nonlinear partial differential equations},
  author={Raissi, Maziar and Perdikaris, Paris and Karniadakis, George E},
  journal={Journal of Computational Physics},
  volume={378},
  pages={686--707},
  year={2019},
  publisher={Elsevier}
}

@article{karniadakis2021physics,
  title={Physics-informed machine learning},
  author={Karniadakis, George Em and Kevrekidis, Ioannis G and Lu, Lu and Perdikaris, Paris and Wang, Sifan and Yang, Liu},
  journal={Nature Reviews Physics},
  volume={3},
  number={6},
  pages={422--440},
  year={2021},
  publisher={Nature Publishing Group UK London}
}

@article{mou2023combining,
  title={Combining stochastic parameterized reduced-order models with machine learning for data assimilation and uncertainty quantification with partial observations},
  author={Mou, Changhong and Smith, Leslie M and Chen, Nan},
  journal={Journal of Advances in Modeling Earth Systems},
  volume={15},
  number={10},
  pages={e2022MS003597},
  year={2023},
  publisher={Wiley Online Library}
}

@book{hughes2003finite,
  title={The finite element method: linear static and dynamic finite element analysis},
  author={Hughes, Thomas JR},
  year={2003},
  publisher={Courier Corporation}
}

@book{evans2022partial,
  title={Partial differential equations},
  author={Evans, Lawrence C},
  volume={19},
  year={2022},
  publisher={American mathematical society}
}

@book{farlow1993partial,
  title={Partial differential equations for scientists and engineers},
  author={Farlow, Stanley J},
  year={1993},
  publisher={Courier Corporation}
}

@book{berselli2006mathematics,
  title={Mathematics of large eddy simulation of turbulent flows},
  author={Berselli, Luigi C and Iliescu, Traian and Layton, William J},
  year={2006},
  publisher={Springer}
}

@book{layton2008introduction,
  title={Introduction to the numerical analysis of incompressible viscous flows},
  author={Layton, William},
  year={2008},
  publisher={SIAM}
}

@book{leveque2007finite,
  title={Finite difference methods for ordinary and partial differential equations: steady-state and time-dependent problems},
  author={LeVeque, Randall J},
  year={2007},
  publisher={SIAM}
}

@article{lu2021learning,
  title={Learning nonlinear operators via DeepONet based on the universal approximation theorem of operators},
  author={Lu, Lu and Jin, Pengzhan and Pang, Guofei and Zhang, Zhongqiang and Karniadakis, George Em},
  journal={Nature machine intelligence},
  volume={3},
  number={3},
  pages={218--229},
  year={2021},
  publisher={Nature Publishing Group UK London}
}

@article{lu2025mopinnenkf,
  title={MoPINNEnKF: Iterative Model Inference using generic-PINN-based ensemble Kalman filter},
  author={Lu, Binghang and Mou, Changhong and Lin, Guang},
  journal={arXiv preprint arXiv:2506.00731},
  year={2025}
}

@article{wang2022respecting,
  title={Respecting causality is all you need for training physics-informed neural networks},
  author={Wang, Sifan and Sankaran, Shyam and Perdikaris, Paris},
  journal={arXiv preprint arXiv:2203.07404},
  year={2022}
}

@article{kovachki2024data,
  title={Data complexity estimates for operator learning},
  author={Kovachki, Nikola B and Lanthaler, Samuel and Mhaskar, Hrushikesh},
  journal={arXiv preprint arXiv:2405.15992},
  year={2024}
}

@article{wang2021learning,
  title={Learning the solution operator of parametric partial differential equations with physics-informed DeepONets},
  author={Wang, Sifan and Wang, Hanwen and Perdikaris, Paris},
  journal={Science advances},
  volume={7},
  number={40},
  pages={eabi8605},
  year={2021},
  publisher={American Association for the Advancement of Science}
}

@article{mou2021data,
  title={Data-driven variational multiscale reduced order models},
  author={Mou, Changhong and Koc, Birgul and San, Omer and Rebholz, Leo G and Iliescu, Traian},
  journal={Computer Methods in Applied Mechanics and Engineering},
  volume={373},
  pages={113470},
  year={2021},
  publisher={Elsevier}
}

@article{chen2020efficient,
  title={Efficient nonlinear optimal smoothing and sampling algorithms for complex turbulent nonlinear dynamical systems with partial observations},
  author={Chen, Nan and Majda, Andrew J},
  journal={Journal of Computational Physics},
  volume={410},
  pages={109381},
  year={2020},
  publisher={Elsevier}
}

@article{mou2023energy,
  title={An energy-based lengthscale for reduced order models of turbulent flows},
  author={Mou, Changhong and Merzari, Elia and San, Omer and Iliescu, Traian},
  journal={Nuclear Engineering and Design},
  volume={412},
  pages={112454},
  year={2023},
  publisher={Elsevier}
}

@article{pou,
	title = {The partition of unity finite element method: {Basic} theory and applications},
	volume = {139},
	issn = {0045-7825},
	shorttitle = {The partition of unity finite element method},
	doi = {10.1016/S0045-7825(96)01087-0},
	journal = {Computer Methods in Applied Mechanics and Engineering},
	author = {Melenk, J. M. and Babuška, I.},
	month = dec,
	year = {1996},
	number = {1},
	pages = {289--314},
}

@article{rbf_pu_pde,
	title = {A {Least} {Squares} {Radial} {Basis} {Function} {Partition} of {Unity} {Method} for {Solving} {PDEs}},
	volume = {39},
	issn = {1064-8275},
	doi = {10.1137/17M1118087},
	journal = {SIAM Journal on Scientific Computing},
	author = {Larsson, Elisabeth and Shcherbakov, Victor and Heryudono, Alfa},
	month = jan,
	year = {2017},
	number = {6},
	pages = {A2538--A2563},
}

@article{rbf_pu_basket,
	title = {Radial basis function partition of unity methods for pricing vanilla basket options},
	volume = {71},
	issn = {0898-1221},
	doi = {10.1016/j.camwa.2015.11.007},
	journal = {Computers \& Mathematics with Applications},
	author = {Shcherbakov, Victor and Larsson, Elisabeth},
	month = jan,
	year = {2016},
	number = {1},
	pages = {185--200},
}

@article{rbf_pu_precond,
	title = {Preconditioning for {Radial} {Basis} {Function} {Partition} of {Unity} {Methods}},
	volume = {67},
	issn = {1573-7691},
	doi = {10.1007/s10915-015-0120-6},
	journal = {Journal of Scientific Computing},
	author = {Heryudono, Alfa and Larsson, Elisabeth and Ramage, Alison and von Sydow, Lina},
	month = jun,
	year = {2016},
	number = {3},
	pages = {1089--1109},
}

@article{rbf_pu_convection,
	title = {A {Radial} {Basis} {Function} {Partition} of {Unity} {Collocation} {Method} for {Convection}–{Diffusion} {Equations} {Arising} in {Financial} {Applications}},
	volume = {64},
	issn = {1573-7691},
	doi = {10.1007/s10915-014-9935-9},
	journal = {Journal of Scientific Computing},
	author = {Safdari-Vaighani, Ali and Heryudono, Alfa and Larsson, Elisabeth},
	month = aug,
	year = {2015},
	number = {2},
	pages = {341--367},
}

@article{SWJCP2018,
	title = {Mesh-free semi-{Lagrangian} methods for transport on a sphere using radial basis functions},
	volume = {366},
	issn = {0021-9991},
	doi = {10.1016/j.jcp.2018.04.007},
	journal = {Journal of Computational Physics},
	author = {Shankar, Varun and Wright, Grady B.},
	month = aug,
	year = {2018},
	pages = {170--190},
}

@article{nattrask_pu,
  title = {A meshfree partition of unity method for elliptic PDEs},
  volume = {284},
  issn = {0021-9991},
  doi = {10.1016/j.jcp.2014.12.026},
  journal = {Journal of Computational Physics},
  author = {Natrask, N.},
  month = feb,
  year = {2015},
  pages = {664--682},
}

@article{goswami2024learning,
  title={Learning stiff chemical kinetics using extended deep neural operators},
  author={Goswami, Somdatta and Jagtap, Ameya D and Babaee, Hessam and Susi, Bryan T and Karniadakis, George Em},
  journal={Computer Methods in Applied Mechanics and Engineering},
  volume={419},
  pages={116674},
  year={2024},
  publisher={Elsevier}
}

@article{liu2021multiscale,
  title={Multiscale DeepONet for nonlinear operators in oscillatory function spaces for building seismic wave responses},
  author={Liu, Lizuo and Cai, Wei},
  journal={arXiv preprint arXiv:2111.04860},
  year={2021}
}

@article{yang2025dd,
  title={DD-DeepONet: Domain decomposition and DeepONet for solving partial differential equations in three application scenarios},
  author={Yang, Bo and Li, Xingquan and Zhao, Jie and Jiang, Ying},
  journal={arXiv preprint arXiv:2508.02717},
  year={2025}
}

@article{chen2025deeponet,
  title={DeepONet-embedded physics-informed neural network for production prediction of multiscale shale matrix--fracture system},
  author={Chen, JiaXuan and Yu, Hao and Li, Bo and Zhang, HouLin and Jin, Xu and Meng, SiWei and Liu, He and Wu, HengAn},
  journal={Physics of Fluids},
  volume={37},
  number={1},
  year={2025},
  publisher={AIP Publishing}
}

@article{mou2025pas,
  title={PAS-Net: Physics-informed Adaptive Scale Deep Operator Network},
  author={Mou, Changhong and Zhang, Yeyu and Zhu, Xuewen and Zhuang, Qiao},
  journal={arXiv preprint arXiv:2511.14925},
  year={2025}
}

@book{hesthaven2016certified,
  title={Certified reduced basis methods for parametrized partial differential equations},
  author={Hesthaven, Jan S and Rozza, Gianluigi and Stamm, Benjamin and others},
  volume={590},
  year={2016},
  publisher={Springer}
}

@article{courant1967partial,
  title={On the partial difference equations of mathematical physics},
  author={Courant, Richard and Friedrichs, Kurt and Lewy, Hans},
  journal={IBM journal of Research and Development},
  volume={11},
  number={2},
  pages={215--234},
  year={1967},
  publisher={IBM}
}

@article{kovachki2024operator,
  title={Operator learning: Algorithms and analysis},
  author={Kovachki, Nikola B and Lanthaler, Samuel and Stuart, Andrew M},
  journal={Handbook of Numerical Analysis},
  volume={25},
  pages={419--467},
  year={2024},
  publisher={Elsevier}
}

@incollection{boulle2024mathematical,
  title={A mathematical guide to operator learning},
  author={Boull{\'e}, Nicolas and Townsend, Alex},
  booktitle={Handbook of Numerical Analysis},
  volume={25},
  pages={83--125},
  year={2024},
  publisher={Elsevier}
}

@article{zhang2022convergence,
  title={Convergence to precipitating quasi-geostrophic equations with phase changes: asymptotics and numerical assessment},
  author={Zhang, Yeyu and Smith, Leslie M and Stechmann, Samuel N},
  journal={Philosophical Transactions of the Royal Society A},
  volume={380},
  number={2226},
  pages={20210030},
  year={2022},
  publisher={The Royal Society}
}

@misc{driscoll2014chebfun,
  title={Chebfun guide},
  author={Driscoll, Tobin A and Hale, Nicholas and Trefethen, Lloyd N},
  year={2014},
  publisher={Pafnuty Publications, Oxford}
}

@article{reitz1981study,
  title={A study of numerical methods for reaction-diffusion equations},
  author={Reitz, Rolf D},
  journal={SIAM Journal on Scientific and Statistical Computing},
  volume={2},
  number={1},
  pages={95--106},
  year={1981},
  publisher={SIAM}
}

@article{chen2024stochastic,
  title={A stochastic precipitating quasi-geostrophic model},
  author={Chen, Nan and Mou, Changhong and Smith, Leslie M and Zhang, Yeyu},
  journal={Physics of Fluids},
  volume={36},
  number={11},
  year={2024},
  publisher={AIP Publishing}
}

@article{zhang2021fast,
  title={Fast-wave averaging with phase changes: asymptotics and application to moist atmospheric dynamics},
  author={Zhang, Yeyu and Smith, Leslie M and Stechmann, Samuel N},
  journal={Journal of Nonlinear Science},
  volume={31},
  number={2},
  pages={38},
  year={2021},
  publisher={Springer}
}

@article{zhang2021effects,
  title={Effects of clouds and phase changes on fast-wave averaging: a numerical assessment},
  author={Zhang, Yeyu and Smith, Leslie M and Stechmann, Samuel N},
  journal={Journal of Fluid Mechanics},
  volume={920},
  pages={A49},
  year={2021},
  publisher={Cambridge University Press}
}

@article{rasheed2020digital,
  title={Digital twin: Values, challenges and enablers from a modeling perspective},
  author={Rasheed, Adil and San, Omer and Kvamsdal, Trond},
  journal={IEEE access},
  volume={8},
  pages={21980--22012},
  year={2020},
  publisher={IEEE}
}

@article{chen2025targeted,
  title={Targeted Digital Twin via Flow Map Learning and Its Application to Fluid Dynamics},
  author={Chen, Qifan and Xu, Zhongshu and Zhang, Jinjin and Xiu, Dongbin},
  journal={arXiv preprint arXiv:2510.07549},
  year={2025}
}

@article{chen2024learning,
  title={Learning stochastic dynamical system via flow map operator},
  author={Chen, Yuan and Xiu, Dongbin},
  journal={Journal of Computational Physics},
  volume={508},
  pages={112984},
  year={2024},
  publisher={Elsevier}
}

\end{document}